%% file: main.tex
\documentclass{article}

\PassOptionsToPackage{numbers,compress}{natbib}
\usepackage[main,final]{neurips_2026}

\usepackage[utf8]{inputenc}
\usepackage[T1]{fontenc}
\usepackage{hyperref}
\usepackage{url}
\usepackage{booktabs}
\usepackage{amsfonts}
\usepackage{amsmath}
\usepackage{amssymb}
\usepackage{nicefrac}
\usepackage{microtype}
\usepackage{xcolor}
\usepackage{graphicx}
\usepackage{makecell}
\usepackage{multirow}
\usepackage{caption}
\usepackage{adjustbox}
\usepackage{float}
\usepackage{placeins}

\graphicspath{{./}{figures/}}
\newcommand{\SpikeSSL}{SpikeSSL}
\title{SpikeSSL: A Universal Spike Inference Framework with Dynamics-Informed State-Space Layers}

\author{%
  Chenghao Yue$^{1,*}$ \quad
  Siming Xing$^{1,*}$ \quad
  Shuran Liu$^{1}$ \quad
  Angran Li$^{2}$ \quad
  Yuanlong Zhang$^{1,\dagger}$ \\
  $^1$School of Life Sciences, Tsinghua University, Beijing, China \\
  $^2$School of Future Information Innovation, Fudan University, Shanghai, China \\
  \texttt{\{yuech26,xsm25,liusr24\}@mails.tsinghua.edu.cn} \\
  \texttt{arli25@m.fudan.edu.cn; yuanlongzhang@tsinghua.edu.cn} \\
  $^*$Equal contribution. $^\dagger$Corresponding author.
}

\begin{document}

\maketitle

\begin{abstract}
Two-photon calcium imaging is a standard tool for recording large neural populations in vivo, yet inferring spikes accurately across the growing diversity of calcium indicators remains an open problem. Existing supervised methods achieve reasonable in-domain accuracy but generalize poorly to unseen indicators, because different indicators induce distinct fluorescence kinetics and signal statistics while existing architectures remain relatively simple generic temporal regressors without dynamics-matched inductive bias. We propose SpikeSSL, a universal spike inference framework whose temporal backbone is a bank of bidirectional IIR state-space layers broadly motivated by calcium dynamics. A multi-modal conditioning encoder maps indicator identity, sampling rate, and trace-level signal statistics into a global conditioning vector that modulates the backbone via Adaptive Layer Normalization, while a heteroscedastic variance head provides calibrated per-frame uncertainty. On a benchmark with five fixed evaluation splits built from 33 public ground-truth datasets, SpikeSSL achieves state-of-the-art performance in both in-domain and zero-shot leave-one-indicator-out settings. We also develop a biophysical simulation pipeline capable of generating paired fluorescence-spike traces with systematically varied kinetic parameters, spike statistics, response nonlinearities, baseline drift, and noise. Using this pipeline, we synthesize approximately 11{,}000 simulated traces. Augmenting training with these data effectively closes the cross-indicator domain gap and improves zero-shot generalization. Code is publicly available at \url{https://github.com/detimage123/SpikeSSL}.
\end{abstract}
\vspace{-8pt}

\input{main_body.tex}

{\small
\bibliographystyle{plainnat}
\bibliography{spike}
}

\clearpage
\appendix
\raggedbottom
\input{original_appendix_content.tex}
\FloatBarrier
\input{rebuttal_appendix_content.tex}

\end{document}

%% file: main_body.tex
\section{Introduction}
Two-photon calcium imaging is a widely used tool for recording neural population activity in vivo \cite{Denk1990, Zipfel2003}. In this modality the observed fluorescence trace is a noisy, nonlinear, and temporally filtered proxy whose relation to the underlying spike train depends heavily on the calcium indicator, optical noise, and cell-type biophysics \cite{Theis2016, Rupprecht2021}. Spike deconvolution, the task of recovering the latent spike train from the observed calcium fluorescence, is therefore an essential computational step. Existing spike inference methods broadly fall into two paradigms. Model-based approaches such as OASIS \cite{Friedrich2017} and MLSpike \cite{Deneux2016} invert explicit generative models of calcium dynamics, but they require indicator-specific kinetic parameters whose optimal values may diverge from measured quantities. Deep learning-based methods such as CASCADE \cite{Rupprecht2021} and ENS\textsuperscript{2} \cite{Zhou2023} learn the inverse mapping from paired fluorescence-electrophysiology recordings and achieve stronger in-domain performance.

\begin{figure}
    \centering
    \includegraphics[width=\linewidth]{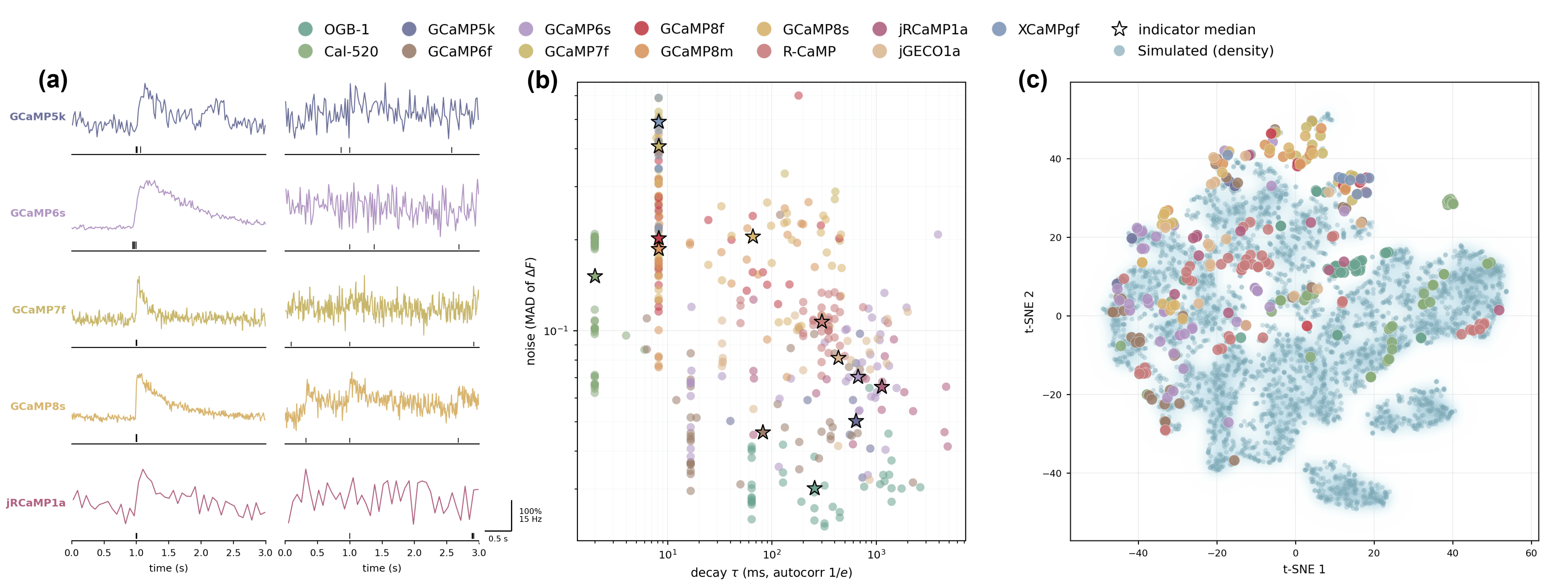}
    \caption{Indicator diversity and coverage gaps. \textbf{(a)} Representative fluorescence and spike snippets show markedly different response shapes across calcium indicators and among related GCaMP variants. \textbf{(b)} Indicator medians separate in kinetic and signal-statistic space, while simulated traces populate sparse regions between real datasets. \textbf{(c)} A t-SNE projection of per-window features shows partially overlapping but systematically displaced indicator domains.}
    \label{fig:intro}
    \vspace{10pt}
\end{figure}

Nevertheless, universal spike inference remains difficult, as models trained on observed indicators can degrade markedly on unseen ones. The central difficulty is that modern calcium indicators induce substantially different fluorescence dynamics. The underlying forward model relating calcium dynamics to fluorescence, parameterized by indicator-specific rise and decay time constants among other factors \cite{Vogelstein2009, Deneux2016, Chen2013, Zhang2023}, varies dramatically across widely used sensors. Figure~\ref{fig:intro}(a,b) illustrates both the differences between different indicators and the data variation within the same indicator. Figure~\ref{fig:intro}(c) further shows that the domain gap between indicator families is large in per-window feature space. At the same time, existing CNN-based supervised models treat spike inference as a generic temporal regression task without inductive bias tailored to calcium dynamics. The forward model itself motivates a structurally matched design, as the first-order linear recurrence governing calcium decay shares the mathematical form of scan operations in deep state-space layers \cite{Gu2022, Gu2024, Smith2023}, with the state-decay coefficient serving as a direct analogue of the indicator time constant. If the network's temporal backbone is built from such recurrences, cross-indicator generalization becomes a more structured problem.

Informed by this observation, we propose SpikeSSL, a universal spike inference framework built on dynamics-informed state-space layers. The temporal backbone consists of bidirectional IIR scan layers spanning a basis of decay rates from fast transient detection to slow baseline tracking. A multi-modal conditioning encoder maps indicator identity, sampling rate, and trace-level signal statistics into a global conditioning vector, which modulates the backbone via Adaptive Layer Normalization as a built-in inductive bias rather than a post-hoc adaptation. In addition, a heteroscedastic variance head provides calibrated per-frame uncertainty and yields wider confidence intervals for out-of-distribution predictions. Although this architecture substantially improves universal inference, the scarcity of paired real recordings still limits cross-indicator generalization. We therefore design a biophysical simulation pipeline that samples spike trains from stochastic point processes and neuron-model dynamics, then passes them through a calcium forward model with varied kinetics, response nonlinearities, depletion, baseline drift, and observation noise. The resulting synthetic traces augment training and successfully bridge the domain gap between real indicator datasets, as illustrated in Figure~\ref{fig:intro}(c).

Our main contributions are as follows.
\begin{itemize}
    \item We propose SpikeSSL, a universal spike inference framework whose bidirectional IIR state-space layers are broadly motivated by calcium indicator decay dynamics. A multi-modal conditioning encoder maps indicator identity, sampling rate, and trace-level signal statistics into a global conditioning vector to modulate the backbone, together with a heteroscedastic variance head for calibrated uncertainty estimation.
    \item On a benchmark with five fixed evaluation splits built from 33 public ground-truth datasets, SpikeSSL achieves state-of-the-art performance in both in-domain and zero-shot LOIO settings, establishing its universal capability across diverse experimental conditions.
    \item We develop a biophysical simulation pipeline that generates synthetic calcium traces with\par\pagebreak\noindent
    systematically varied kinetic parameters and firing statistics. We synthesize approximately 11{,}000 simulated traces covering a wide range of indicator kinetic regimes, effectively closing the cross-indicator gap in zero-shot transfer.
\end{itemize}

\section{Related Work}

\subsection{Spike Inference from Calcium Imaging}
Spike inference methods divide into model-based and data-driven approaches. Model-based approaches invert explicit calcium dynamics models, including linear deconvolution \cite{Yaksi2006}, Bayesian approaches \cite{Vogelstein2009, Vogelstein2010, Pnevmatikakis2013}, the biophysical MLSpike model \cite{Deneux2016}, and the OASIS active-set method \cite{Friedrich2017}. These methods require indicator-dependent kinetic parameters whose optimal values often diverge from biophysically measured quantities \cite{Rupprecht2025a}. Data-driven methods, from early SVM approaches \cite{Sasaki2008} to the deep architectures advanced through the Spikefinder benchmark \cite{Berens2018}, have converged on end-to-end learned representations. CASCADE \cite{Rupprecht2021} provides noise-optimized training across 27 datasets and expands coverage via calibrated noise augmentation, but does not vary kinetic parameters across indicators. ENS\textsuperscript{2} \cite{Zhou2023} offers a compact 1D U-Net that achieves competitive accuracy with fewer parameters. Existing supervised methods rely on largely generic temporal backbones, and their performance degrades on unseen indicators. The Spikefinder benchmark \cite{Berens2018} underscored this pattern, showing that improvements on one indicator rarely transferred to others, and coverage of the CASCADE database \cite{Rupprecht2021} remains uneven, with GCaMP8 variants and red-shifted sensors \cite{Zhang2023, Rupprecht2025a} underrepresented relative to established GECIs, leaving models vulnerable to silent degradation on newly developed indicators.

\subsection{Cross-Domain Generalization}
Cross-domain generalization studies how models trained on multiple source domains transfer to unseen targets without target-domain labels \cite{Muandet2013, Gulrajani2021, Zhou2022DG}. In spike inference, domains are naturally defined by calcium indicators, and the challenge is accurate inference on held-out indicators with kinetics and statistics unseen during training. This problem is exacerbated by the limited diversity of paired ground-truth recordings. Although the database assembled by \citet{Rupprecht2021} broadened coverage across indicators and experimental settings, and has since been extended with recordings for newly developed indicators and additional tissue types \cite{Rupprecht2025a, Rupprecht2025b}, important gaps remain in indicator kinetic space. Recent work shows that CASCADE trained on one domain, such as cortex or GCaMP6-dominated data, transfers suboptimally to spinal cord recordings or GCaMP8 indicators \cite{Rupprecht2025b, Rupprecht2025a}, highlighting the need for stronger cross-indicator generalization. Simulation-based augmentation offers a complementary strategy, where a calibrated forward model can generate paired fluorescence--spike traces spanning a systematic grid of kinetic parameters, extending the training distribution into regimes where real recordings are absent. This direction has received limited exploration in spike inference despite its potential to bridge coverage gaps without additional electrophysiology experiments.

\section{Methodology}

In this section, we present SpikeSSL, a universal spike inference framework with dynamics-informed state-space layers that integrates a bidirectional IIR backbone, multi-modal conditioning, and output heads for spike rate and uncertainty estimation. We then describe the biophysical simulation pipeline used for cross-indicator data augmentation.

\subsection{Overall Architecture}

SpikeSSL is a dynamics-informed network for spike inference whose architecture draws broad motivation from the biophysics of calcium indicators. The model takes a univariate fluorescence trace $y \in \mathbb{R}^T$ together with indicator identity and sampling rate, and produces a per-frame spike rate $\hat{s} \in \mathbb{R}_{\geq 0}^T$ along with calibrated uncertainty $\hat{\sigma}^2 \in \mathbb{R}_{>0}^T$. The trace is first projected to model dimension $D$ via a $1 \times 1$ convolution, while a conditioning encoder maps indicator identity, sampling rate, and trace-level signal statistics through dedicated branches into a global conditioning vector $c \in \mathbb{R}^{64}$.

The projected trace then passes through $N$ stacked blocks, each containing a bidirectional IIR scan with AdaLN conditioning, a Temporal Mixer, and a SwiGLU feed-forward network (FFN).
\begin{figure}
    \centering
    \includegraphics[width=1\linewidth]{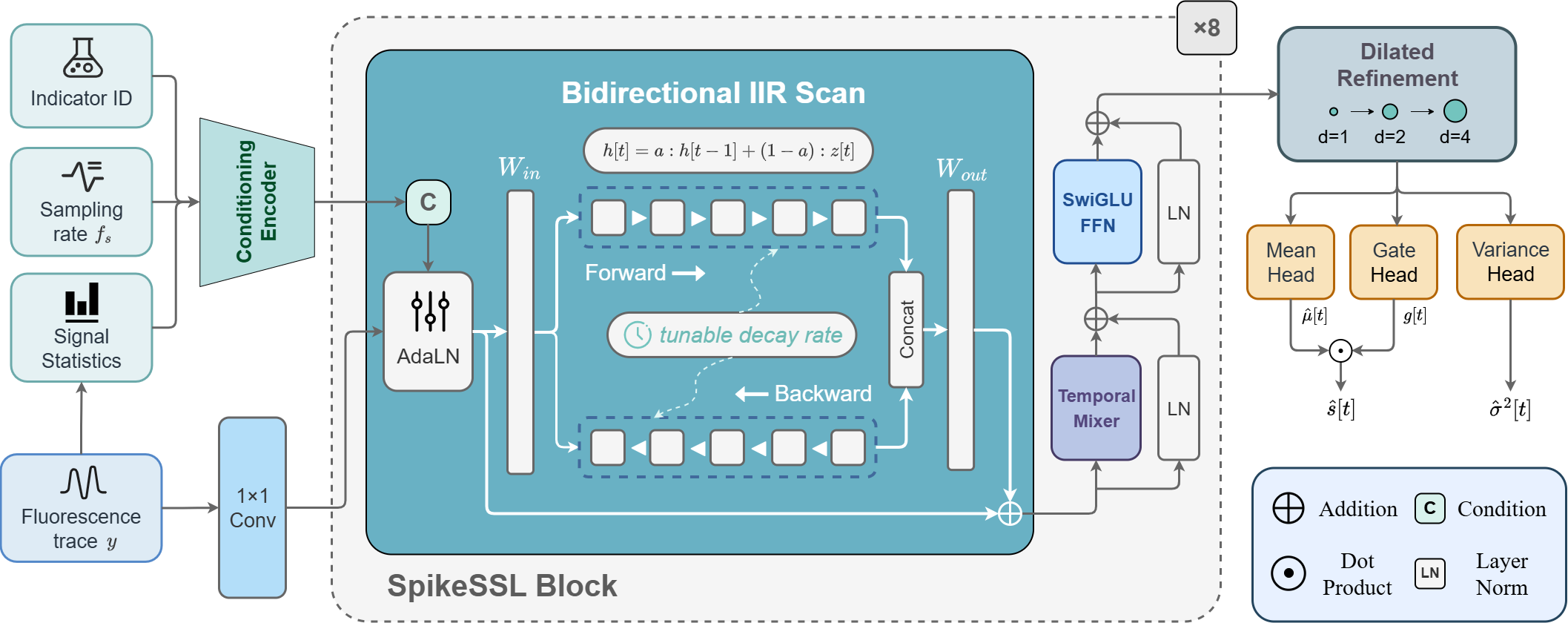}
    \caption{Architecture of SpikeSSL. A fluorescence trace is projected by a pointwise convolution and processed by a stack of SpikeSSL blocks. Each block contains a bidirectional IIR scan with learnable decay rates that capture the exponential decay dynamics of calcium indicators, modulated by AdaLN conditioning on indicator identity, sampling rate, and trace-level signal statistics, followed by a Temporal Mixer for local temporal mixing and a SwiGLU FFN for channel mixing. A dilated refinement module aggregates multi-scale temporal context. Three parallel output heads produce a spike rate estimate, a binary spiking gate, and a per-frame uncertainty estimate.}
    \label{fig:main}
    \vspace{12pt}
\end{figure}
After the stack, a dilated refinement module of three residual convolutional layers with dilation schedule 1, 2, and 4 expands the receptive field from local to medium-range context, complementing the long-range multi-scale decay basis encoded by the IIR layers.

Three output heads operate on the refined features. The mean head predicts the non-negative spike rate $\hat{\mu}[t]$ via a convolutional projection followed by ReLU. A sigmoid gate head $g[t] = \sigma(w_g^\top h[t])$ classifies each frame as spiking or silent, is detached from the mean head's gradient during training, and is trained with a separate binary cross-entropy loss, preventing co-adaptation between the two heads. At inference, $\hat{s}[t] = \hat{\mu}[t]$ if $g[t] > 0.5$ and $\hat{s}[t] = 0$ otherwise, eliminating low-amplitude noise in silent periods. A variance head produces per-frame uncertainty estimates $\hat{\sigma}^2[t]$ as described in Section~\ref{sec:uncertainty}.

\subsection{Dynamics-Informed IIR Backbone}

The central design principle of SpikeSSL is to ground the network's temporal processing in the dynamics of calcium indicators. Simple first-order exponential decay is among the most common and general models of calcium dynamics, and although the underlying biophysics can be substantially more complex, this recurrence still provides a useful inductive bias.
The IIR scan used in deep state-space layers \cite{Gu2022, Gu2024, Smith2023, Gupta2022, Fu2023, Poli2023} follows a structurally similar recurrence,
\begin{equation}\label{eq:iir}
    h[t] = \alpha \cdot h[t{-}1] + (1 - \alpha) \cdot z[t],
\end{equation}
where $\alpha = \sigma(\alpha_\mathrm{logit})$ is a per-channel decay rate parameterized via the sigmoid function, and $z$ is the input projected from the trace embedding. The resemblance to exponential calcium decay motivates initializing and conditioning the decay rates to span the physiologically relevant range. By initializing the base logits as $\alpha_\mathrm{logit} \sim \mathrm{linspace}(0, 4)$, the model spans a multi-scale bank of decay rates from fast time constants of a few frames to slow trends unfolding over hundreds of frames. This range deliberately covers the full spread of indicator kinetics observed in the benchmark, providing a structured inductive bias that reflects the underlying biophysics rather than leaving the network to discover relevant temporal scales from data statistics alone.

To capture both causal context from past fluorescence and anti-causal context from future decay shape, SpikeSSL employs bidirectional IIR scans. Each layer computes a forward scan $h^\mathrm{fwd}$ and a time-reversed backward scan $h^\mathrm{bwd}$, concatenates them along the channel dimension, and projects back to the model dimension,
\begin{equation}
    h = W_\mathrm{out} \, [\, h^\mathrm{fwd} \;\|\; h^\mathrm{bwd} \,] + b_\mathrm{out}.
\end{equation}

\subsection{Multi-Modal Conditioning}

Different indicators exhibit different kinetics, and SpikeSSL must adapt its decay rates accordingly. We achieve this through a multi-modal conditioning encoder, Adaptive Layer Normalization (AdaLN) \cite{Peebles2023}, and per-sample decay rate modulation.

The conditioning vector $c \in \mathbb{R}^{d_c}$ fuses three sources of information,
\begin{equation}
    c = \mathrm{MLP}\big(\, e_\mathrm{ind} \;\|\; f_\mathrm{fs} \;\|\; f_\mathrm{stats} \,\big),
\end{equation}
where $e_\mathrm{ind}$ is a learned embedding of the discrete indicator identity, $f_\mathrm{fs} = \mathrm{MLP}(\log f_s)$ encodes the sampling rate, and $f_\mathrm{stats}$ encodes trace-level signal statistics. Each source is independently projected to a shared dimension before concatenation and fusion.

The signal statistics component extracts a 7-dimensional summary vector from the input trace,
\begin{equation}
    f_\mathrm{stats} = \big[\, \hat{\sigma}_y, \;\hat{\gamma}_y, \;\hat{\kappa}_y, \;\rho_1, \;\rho_3, \;\rho_5, \;\rho_{10} \,\big],
\end{equation}
comprising the standard deviation, skewness, excess kurtosis, and normalized autocorrelation at lags 1, 3, 5, and 10. These statistics are computed deterministically with no learnable parameters and serve as an implicit proxy for the indicator's SNR and kinetic properties. Slowly decaying indicators produce high autocorrelation, while noisy or fast indicators exhibit low autocorrelation and high kurtosis. This module enables SpikeSSL to adapt its temporal dynamics even when the indicator type is not specified.

The conditioning vector modulates the IIR sub-block via AdaLN with zero-initialized projection,
\begin{equation}
    \mathrm{AdaLN}(x, c) = \mathrm{LN}(x) \odot (1 + \gamma(c)) + \beta(c),
\end{equation}
where $(\gamma, \beta) = W_c \, c$ is a linear projection initialized to zero, so that at initialization the model reduces to standard Layer Normalization. AdaLN is applied only to the temporal IIR sub-block, while the subsequent convolution and feed-forward sub-blocks use standard normalization. This concentrates the indicator-conditioned modulation on the component that directly encodes decay dynamics.

In addition to AdaLN, the conditioning vector directly shifts the IIR decay logits,
\begin{equation}
    \alpha_\mathrm{eff} = \sigma\big(\alpha_\mathrm{logit} + W_\alpha \, c\big),
\end{equation}
where $W_\alpha$ is zero-initialized. This allows each sample's effective decay rates to be continuously modulated by its indicator identity and signal properties, enhancing the model's ability to adapt to diverse kinetic regimes.

\subsection{Heteroscedastic Uncertainty Estimation}\label{sec:uncertainty}

SpikeSSL produces a per-frame variance estimate $\hat{\sigma}^2[t]$ alongside the spike rate prediction. The variance head consists of a single convolutional projection from the shared backbone features plus a per-indicator learned bias,
\begin{equation}
    \log \hat{\sigma}^2[t] = w_v^\top \, h[t] + b_\mathrm{ind},
\end{equation}
where $b_\mathrm{ind}$ is an indicator-specific scalar bias. Both the projection and bias are zero-initialized. A key emergent property is that the variance output serves as an automatic out-of-distribution detector, as in-domain predictions yield tight confidence intervals while zero-shot predictions on unseen indicators produce inflated variance.

\begin{figure}
    \centering
    \includegraphics[width=0.97\linewidth]{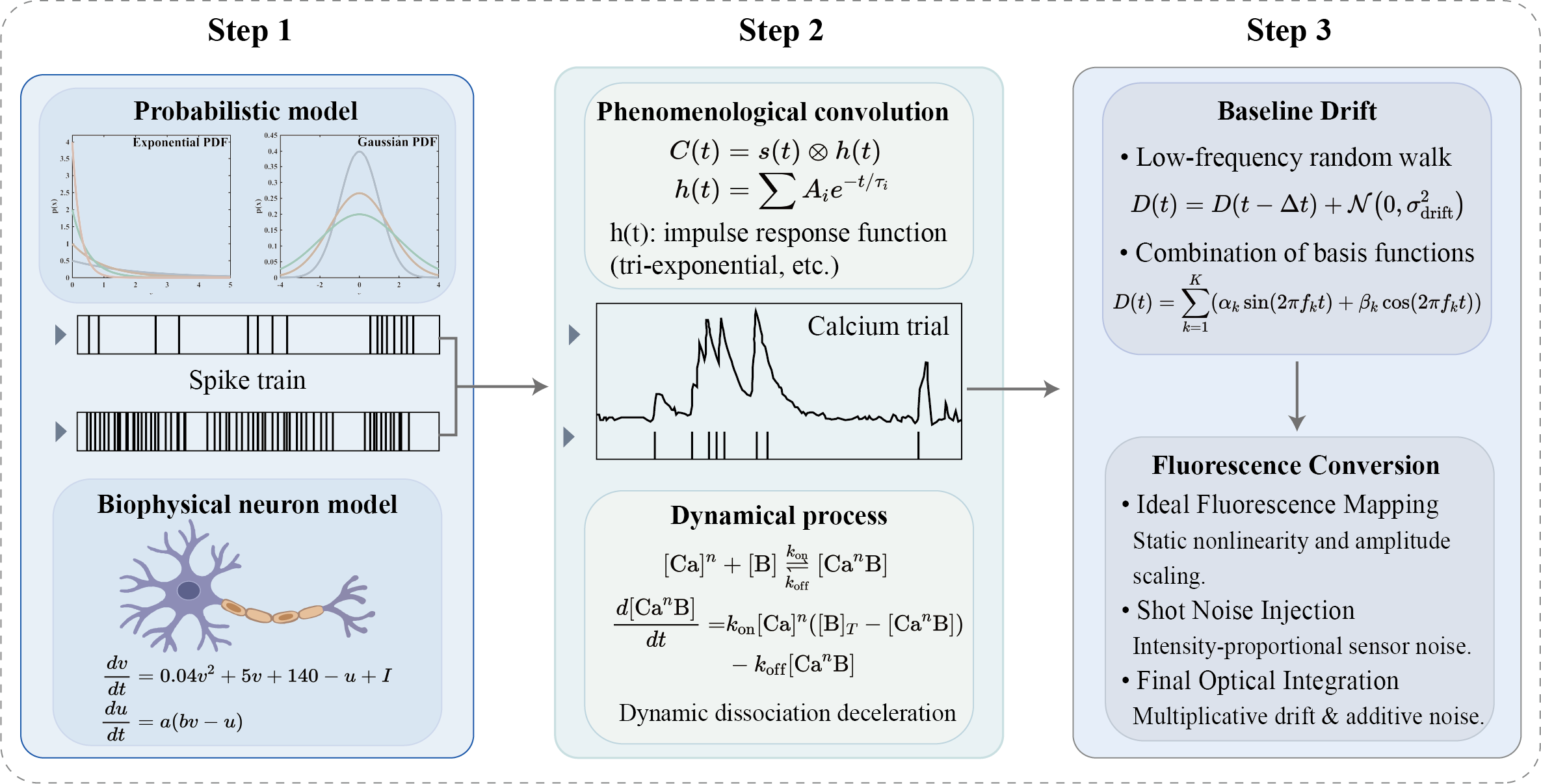}
    \caption{Hierarchical Workflow of the Calcium Data Simulator. }
    % The framework generates realistic calcium fluorescence data through three sequential steps. Step 1 simulates diverse neuronal spike trains using either stochastic distributions or biophysical equations to replicate a wide range of physiological firing patterns. Step 2 calculates the underlying calcium dynamics via a dual-engine architecture: a phenomenological convolution model and an extended kinetic model. Step 3 encodes the calcium-bound fraction into realistic optical signals, followed by the integration of low-frequency baseline drift and signal-dependent shot noise to yield the final simulated observations.
    \label{fig:sim_pipeline}
\end{figure}

\subsection{Biophysical Simulation Data Generation}\label{sec:sim}

To populate underrepresented regimes while preserving realistic action-potential-driven fluorescence, we augment training with simulated calcium traces generated by an enhanced MLSpike-based forward model \cite{Deneux2016}. The simulator complements real paired recordings by expanding kinetic coverage rather than replacing empirical data, and samples parameters across ranges covering observed indicators and plausible unseen sensors. As illustrated in Fig.~\ref{fig:sim_pipeline}, the simulation pipeline contains three components: spike train synthesis, calcium transient generation, and observation noise modeling. For each trial, we sample a recording duration and target firing rate, then generate spike trains with two complementary strategies. One strategy draws spike counts and inter-spike intervals from stochastic point-process models, allowing sparse, irregular, quasi-periodic, and burst-like firing. The other integrates the Izhikevich neuron model \cite{Izhikevich2003} to produce spikes from continuous membrane potential dynamics. Together, these strategies provide a broad repertoire of neuronal activity regimes before the fluorescence forward model is applied.

The generated spikes are then passed through a modified calcium simulator. Compared with standard MLSpike, our simulator introduces stochastic amplitude jitter, a recovery variable that captures depletion and interval dependent response scaling, and indicator specific cooperative binding kinetics. Finally, we form observable fluorescence by adding slow baseline drift and Gaussian measurement noise to the modeled calcium response, producing traces with variability in baseline stability and signal quality. These mechanisms broaden the range of firing patterns, response amplitudes, noise conditions, and indicator dynamics beyond the real training set alone, helping bridge held out kinetic regimes during zero shot transfer. \textit{More details are provided in the supplementary material.}

\subsection{Training Objective}

The total loss combines three components, namely a heteroscedastic reconstruction loss, structural penalty terms, and a gate detection loss,
\begin{equation}
    \mathcal{L} = \mathcal{L}_\mathrm{recon} + \mathcal{L}_\mathrm{struct} + \lambda_g \,\mathcal{L}_\mathrm{gate}.
\end{equation}

The reconstruction loss follows the $\beta$-NLL formulation \cite{Seitzer2022}, which precision-weights the per-frame Huber loss by the predicted variance,
\begin{equation}
    \mathcal{L}_\mathrm{recon} = \frac{1}{\hat{\sigma}^2} \cdot \mathrm{Huber}_\delta(\hat{s}, s) + \tfrac{1}{2} \log \hat{\sigma}^2.
\end{equation}
High-confidence frames receive amplified gradients while noisy frames are down-weighted, and the entropy term $\tfrac{1}{2}\log\hat{\sigma}^2$ prevents the degenerate solution $\hat{\sigma}^2 \to \infty$.

The structural penalties impose three asymmetric constraints that are not attenuated by the variance estimate, namely an over-prediction penalty that penalizes $\hat{s}[t] > s[t]$ more heavily than under-prediction, a peak under-prediction penalty that pulls up missed high-amplitude spike events, and a mean bias penalty that prevents systematic offset between predicted and true mean spike rates,
\begin{equation}
    \mathcal{L}_\mathrm{struct} = \beta_\mathrm{over} \cdot \mathcal{L}_\mathrm{over} + \beta_\mathrm{peak} \cdot \mathcal{L}_\mathrm{peak} + \lambda_\mu \cdot \mathcal{L}_\mathrm{bias}.
\end{equation}

\begin{table*}[!t]
    \setlength{\tabcolsep}{2.5pt}
    \renewcommand\arraystretch{1.05}
    \centering
    \caption{In-domain resluts. All methods are trained and evaluated on the full ground-truth database with per-dataset train and test splits. \textbf{Bold} / \underline{Underline} indicate the 1st / 2nd best results.}
    \label{tab:indomain}
    \resizebox{\linewidth}{!}{
    \begin{tabular}{@{}l ccc ccc ccc ccc ccc c c @{}}
        \toprule
        \multirow{2}{*}{Method}
          & \multicolumn{3}{c}{V1-OGB}
          & \multicolumn{3}{c}{V1-GCaMP6s}
          & \multicolumn{3}{c}{V1-GCaMP8}
          & \multicolumn{3}{c}{Other-RCaMP}
          & \multicolumn{3}{c}{SC-GCaMP6s}
          & Mean
          & \multirow{2}{*}{Params} \\
        \cmidrule(lr){2-4} \cmidrule(lr){5-7} \cmidrule(lr){8-10} \cmidrule(lr){11-13} \cmidrule(lr){14-16} \cmidrule(lr){17-17}
          & Corr\,$\uparrow$ & Err\,$\downarrow$ & Bias\,$\downarrow$
          & Corr\,$\uparrow$ & Err\,$\downarrow$ & Bias\,$\downarrow$
          & Corr\,$\uparrow$ & Err\,$\downarrow$ & Bias\,$\downarrow$
          & Corr\,$\uparrow$ & Err\,$\downarrow$ & Bias\,$\downarrow$
          & Corr\,$\uparrow$ & Err\,$\downarrow$ & Bias\,$\downarrow$
          & Corr\,$\uparrow$ & \\
        \midrule
        OASIS
          & 0.298 & \underline{1.157} & $-$0.481
          & 0.332 & \underline{0.973} & $-$0.883
          & 0.493 & \underline{0.878} & \underline{$-$0.659}
          & 0.504 & \underline{1.054} & $-$0.508
          & \underline{0.365} & \underline{0.975} & $-$0.964
          & 0.398 
          & --- \\
        MLSpike
          & 0.363 & 4.379 & \hphantom{$-$}3.244
          & 0.224 & 1.017 & $-$0.928
          & 0.558 & 0.935 & $-$0.887
          & 0.354 & 1.369 & $-$0.761
          & 0.307 & 1.005 & $-$0.919
          & 0.361 
          & --- \\
        CASCADE~\cite{Rupprecht2021}
          & 0.613 & 1.117 & \underline{$-$0.250}
          & \underline{0.361} & 1.718 & \textbf{\hphantom{$-$}0.520}
          & \underline{0.825} & 3.246 & \hphantom{$-$}3.108
          & 0.675 & 1.060 & \underline{$-$0.388}
          & 0.149 & 1.005 & \underline{$-$0.639}
          & 0.524
          & 34K \\
        ENS\textsuperscript{2}~\cite{Zhou2023}
          & \underline{0.641} & 1.032 & $-$0.368
          & 0.358 & 2.265 & \hphantom{$-$}1.354
          & 0.775 & 1.827 & \hphantom{$-$}1.596
          & \underline{0.695} & \textbf{0.884} & $-$0.509
          & 0.256 & 1.116 & \textbf{$-$0.154}
          & \underline{0.545}
          & 145K \\
        SpikeSSL
          & \textbf{0.743} & \textbf{0.810} & \textbf{$-$0.067}
          & \textbf{0.579} & \textbf{0.945} & \underline{$-$0.741}
          & \textbf{0.915} & \textbf{0.565} & \textbf{\hphantom{$-$}0.017}
          & \textbf{0.731} & 1.269 & \textbf{\hphantom{$-$}0.234}
          & \textbf{0.736} & \textbf{0.764} & $-$0.704
          & \textbf{0.741}
          & 3.9M \\
        \bottomrule
    \end{tabular}}
\end{table*}

\begin{table*}[!t]
    \setlength{\tabcolsep}{2.5pt}
    \renewcommand\arraystretch{1.05}
    \centering    \caption{Zero-shot LOIO results. Each column group holds out one of the five evaluation splits during training and evaluates on it without target-domain data. \textbf{Bold} / \underline{Underline} indicate the 1st / 2nd best results.}
    \label{tab:loio}
    \resizebox{\linewidth}{!}{
    \begin{tabular}{@{}l ccc ccc ccc ccc ccc c @{}}
        \toprule
        \multirow{2}{*}{Method}
          & \multicolumn{3}{c}{V1-OGB}
          & \multicolumn{3}{c}{V1-GCaMP6s}
          & \multicolumn{3}{c}{V1-GCaMP8}
          & \multicolumn{3}{c}{Other-RCaMP}
          & \multicolumn{3}{c}{SC-GCaMP6s}
          & Mean \\
        \cmidrule(lr){2-4} \cmidrule(lr){5-7} \cmidrule(lr){8-10} \cmidrule(lr){11-13} \cmidrule(lr){14-16} \cmidrule(lr){17-17}
          & Corr\,$\uparrow$ & Err\,$\downarrow$ & Bias\,$\downarrow$
          & Corr\,$\uparrow$ & Err\,$\downarrow$ & Bias\,$\downarrow$
          & Corr\,$\uparrow$ & Err\,$\downarrow$ & Bias\,$\downarrow$
          & Corr\,$\uparrow$ & Err\,$\downarrow$ & Bias\,$\downarrow$
          & Corr\,$\uparrow$ & Err\,$\downarrow$ & Bias\,$\downarrow$
          & Corr\,$\uparrow$ \\
        \midrule
        OASIS
          & 0.298 & \underline{1.157} & $-$0.481
          & \underline{0.332} & \underline{0.973} & \underline{$-$0.883}
          & 0.493 & \underline{0.878} & \underline{$-$0.659}
          & 0.504 & \underline{1.054} & $-$0.508
          & \underline{0.365} & \underline{0.975} & $-$0.964
          & 0.398 \\
        MLSpike
          & 0.363 & 4.379 & \hphantom{$-$}3.244
          & 0.224 & 1.017 & $-$0.928
          & 0.558 & 0.935 & $-$0.887
          & 0.354 & 1.369 & $-$0.761
          & 0.307 & 1.005 & $-$0.919
          & 0.361 \\
        CASCADE
          & 0.560 & 1.208 & \textbf{$-$0.144}
          & 0.304 & 2.571 & 1.614
          & \underline{0.757} & 3.723 & 3.602
          & 0.551 & 1.086 & \underline{$-$0.300}
          & 0.126 & 1.083 & \underline{$-$0.442}
          & 0.460 \\
        ENS\textsuperscript{2}
          & \underline{0.574} & \textbf{1.005} & \underline{$-$0.366}
          & 0.310 & 3.401 & 2.654
          & 0.716 & 2.038 & 1.715
          & \underline{0.635} & \underline{0.966} & \underline{$-$0.454}
          & 0.178 & 1.092 & \textbf{$-$0.391}
          & \underline{0.483} \\
        SpikeSSL
          & \textbf{0.650} & 1.979 & \hphantom{$-$}0.675
          & \textbf{0.509} & \textbf{0.944} & \textbf{$-$0.677}
          & \textbf{0.853} & \textbf{0.513} & \textbf{$-$0.091}
          & \textbf{0.717} & 1.172 & \hphantom{$-$}0.606
          & \textbf{0.542} & \textbf{0.911} & $-$0.892
          & \textbf{0.654} \\
        \bottomrule
    \end{tabular}}
\end{table*}

The gate loss trains the binary spike-or-silent classifier via cross-entropy, $\mathcal{L}_\mathrm{gate} = \mathrm{BCE}(g[t], \,\mathbf{1}[s[t]>\epsilon])$, with gradients detached from the mean head to prevent co-adaptation.

\section{Experiments}

\subsection{Datasets}\label{sec:datasets}

We evaluate on the ground-truth database assembled by \citet{Rupprecht2021}, comprising 33 datasets with 1{,}960 paired neuron recordings spanning 11 calcium indicators, two species (mouse and zebrafish), multiple brain regions (V1, CA3, olfactory bulb, telencephalon) and spinal cord, with both excitatory neurons and inhibitory interneurons (PV, SST, VIP). The corpus covers chemical dyes (OGB-1, Cal-520), legacy GECIs (GCaMP5k, 6f, 6s, 7f \cite{Chen2013}), next-generation GCaMP8 variants (8f, 8m, 8s \cite{Zhang2023}), and red-shifted indicators (R-CaMP, jRCaMP1a, jGECO1a). Recent spinal cord recordings \cite{Rupprecht2025b} introduce a major tissue-type shift from conventional cortical preparations. We additionally augment training with biophysically simulated calcium traces stored in multiple generated batches (Section~\ref{sec:sim} and Appendix~\ref{app:sim_details}) used exclusively during training.

\subsection{Benchmark and Metrics}\label{sec:benchmark}

We compare SpikeSSL against four baselines spanning the major paradigms, namely OASIS \cite{Friedrich2017}, MLSpike \cite{Deneux2016}, CASCADE \cite{Rupprecht2021}, and ENS\textsuperscript{2} \cite{Zhou2023}. The underlying recordings are publicly available, and we evaluate them under five fixed evaluation splits. In the \emph{in-domain} setting, all splits are included during training with per-dataset train and test subsets. In the LOIO setting, five representative experimental conditions, namely V1-OGB, V1-GCaMP6s, V1-GCaMP8, Other-RCaMP, and SC-GCaMP6s, are each held out in turn and evaluated without any target-domain training data. Across this benchmark, indicator decay time constants span from ${\sim}50$\,ms for GCaMP8f to ${\sim}600$\,ms for GCaMP6s, creating fundamentally different inverse problems for each condition.

We evaluate all methods using Pearson correlation between the predicted and ground-truth spike rate \cite{Theis2016, Rupprecht2021, Zhou2023}, computed per neuron and averaged across neurons within each dataset. We additionally report normalized error and mean bias (signed difference in temporal means) to assess systematic over- or under-prediction.

\begin{table*}[!t]
    \setlength{\tabcolsep}{2.5pt}
    \renewcommand\arraystretch{1.05}
    \centering
    \caption{LOIO results with simulation augmentation (``+ Sim.''). Training is augmented with biophysically simulated calcium traces. \textbf{Bold} / \underline{Underline} indicate the 1st / 2nd best results.}
    \label{tab:loio_sim}
    \resizebox{\linewidth}{!}{
    \begin{tabular}{@{}l ccc ccc ccc ccc ccc c @{}}
        \toprule
        \multirow{2}{*}{Method}
          & \multicolumn{3}{c}{V1-OGB}
          & \multicolumn{3}{c}{V1-GCaMP6s}
          & \multicolumn{3}{c}{V1-GCaMP8}
          & \multicolumn{3}{c}{Other-RCaMP}
          & \multicolumn{3}{c}{SC-GCaMP6s}
          & Mean \\
        \cmidrule(lr){2-4} \cmidrule(lr){5-7} \cmidrule(lr){8-10} \cmidrule(lr){11-13} \cmidrule(lr){14-16} \cmidrule(lr){17-17}
          & Corr\,$\uparrow$ & Err\,$\downarrow$ & Bias\,$\downarrow$
          & Corr\,$\uparrow$ & Err\,$\downarrow$ & Bias\,$\downarrow$
          & Corr\,$\uparrow$ & Err\,$\downarrow$ & Bias\,$\downarrow$
          & Corr\,$\uparrow$ & Err\,$\downarrow$ & Bias\,$\downarrow$
          & Corr\,$\uparrow$ & Err\,$\downarrow$ & Bias\,$\downarrow$
          & Corr\,$\uparrow$ \\
        \midrule
        CASCADE + Sim.
          & \underline{0.552} & 2.418 & \hphantom{$-$}1.423
          & 0.439 & \underline{1.522} & \textbf{\hphantom{$-$}0.691}
          & 0.716 & \underline{1.408} & \underline{\hphantom{$-$}1.018}
          & \underline{0.642} & \underline{1.421} & \underline{\hphantom{$-$}0.614}
          & 0.405 & 3.206 & \hphantom{$-$}2.084
          & 0.551 \\
        ENS\textsuperscript{2} + Sim.
          & 0.486 & \underline{1.634} & \underline{\hphantom{$-$}0.592}
          & \underline{0.447} & 2.146 & \hphantom{$-$}1.418
          & \underline{0.796} & 1.882 & \hphantom{$-$}1.487
          & 0.588 & 1.764 & \hphantom{$-$}0.756
          & \underline{0.461} & \underline{2.291} & \underline{\hphantom{$-$}1.054}
          & \underline{0.556} \\
        SpikeSSL + Sim.
          & \textbf{0.697} & \textbf{1.441} & \textbf{\hphantom{$-$}0.419}
          & \textbf{0.533} & \textbf{0.931} & \underline{$-$0.697}
          & \textbf{0.902} & \textbf{0.523} & \textbf{\hphantom{$-$}0.175}
          & \textbf{0.708} & \textbf{0.849} & \textbf{$-$0.505}
          & \textbf{0.678} & \textbf{0.771} & \textbf{$-$0.653}
          & \textbf{0.704} \\
        \bottomrule
    \end{tabular}}
\end{table*}

\begin{figure*}[!t]
    \centering
    \includegraphics[width=\linewidth]{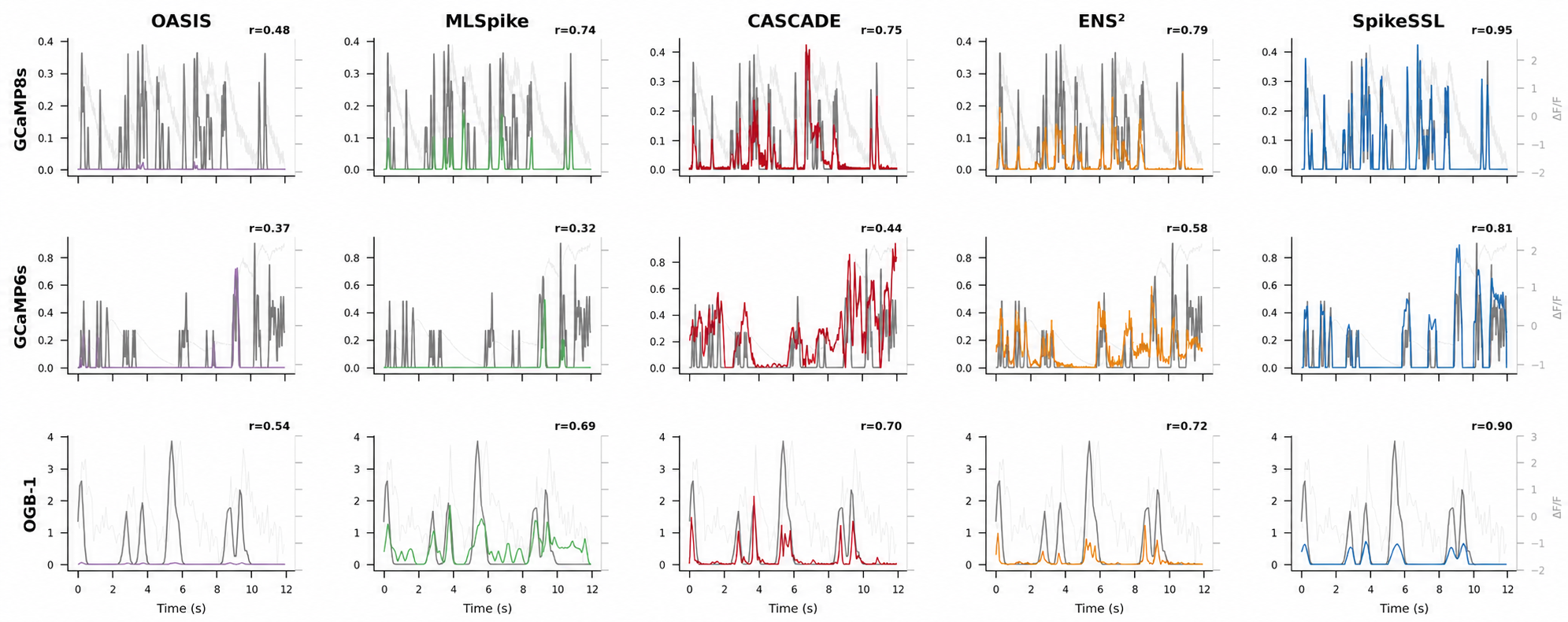}
    \caption{Qualitative LOIO comparison on GCaMP8s, GCaMP6s, and OGB-1 test neurons. Each column shows a different method, and the grey curve denotes the ground-truth smoothed spike rate.}
    \label{fig:vis_loio}
\end{figure*}

\subsection{Implementation Details}

SpikeSSL uses 8 stacked blocks with model dimension $D{=}192$ and conditioning dimension $d_c{=}64$. We train with Adam  using an initial learning rate of $10^{-3}$. The model converges within 20 epochs. For the loss hyperparameters, we set the over-prediction penalty $\beta_\mathrm{over}{=}3.0$, peak under-prediction weight $\beta_\mathrm{under}{=}0.3$, Huber threshold $\delta{=}0.1$, and mean bias weight $\lambda_\mu{=}0.5$. The gate loss weight $\lambda_g{=}5.0$ is set deliberately large to suppress false-positive spike predictions in silent periods.

\subsection{In-Domain Results}

Table~\ref{tab:indomain} reports in-domain performance with all evaluation splits included during training. SpikeSSL achieves state-of-the-art performance, substantially outperforming all baselines. These results demonstrate that the dynamics-informed IIR backbone and multi-modal conditioning provide strong inductive bias even in the fully supervised setting where all indicator types are observed during training. The consistently lower error and bias values confirm that SpikeSSL produces well-calibrated spike rate estimates across the full diversity of indicators.

\subsection{Zero-Shot LOIO Results}

Under the LOIO protocol defined in Section~\ref{sec:benchmark}, cross-indicator generalization remains difficult for all methods, but SpikeSSL achieves the highest mean correlation across all five held-out evaluation splits in Table~\ref{tab:loio}. This result indicates that the IIR backbone and indicator-aware conditioning provide a meaningful structural advantage under domain shift. When training is augmented with biophysically simulated calcium traces, zero-shot performance improves further across all held-out splits in Table~\ref{tab:loio_sim}, with particularly large gains on conditions whose kinetics are sparsely covered by the real training set. Figure~\ref{fig:vis_loio} shows that SpikeSSL preserves sharper spike structure than the classical baselines on representative unseen indicators.

\begin{figure*}[t]
\centering
\begin{minipage}[t]{0.23\textwidth}
    \vspace{0pt}
    \centering
    \setlength{\tabcolsep}{2pt}
    \captionof{table}{Ablation study under LOIO. Mean correlation is averaged over all five held-out evaluation splits.}
    \label{tab:ablation}
    \scriptsize
    \renewcommand\arraystretch{1.1}
    \begin{tabular}{@{}cc@{}}
        \toprule
        Method & Mean Corr ($\uparrow$) \\
        \midrule
        V1 ($-$ AdaLN)              & 0.601 \\
        V2 ($-$ Stats)              & 0.623 \\
        V3 ($-$ Bi-IIR)             & 0.594 \\
        V4 ($-$ Var.)               & 0.608 \\
        V5 ($-$ Gate)               & 0.621 \\
        \midrule
        \textbf{SpikeSSL}               & \textbf{0.654} \\
        \bottomrule
    \end{tabular}
\end{minipage}
\hfill
\begin{minipage}[t]{0.73\textwidth}
    \vspace{0pt}
    \centering
    \includegraphics[width=\textwidth]{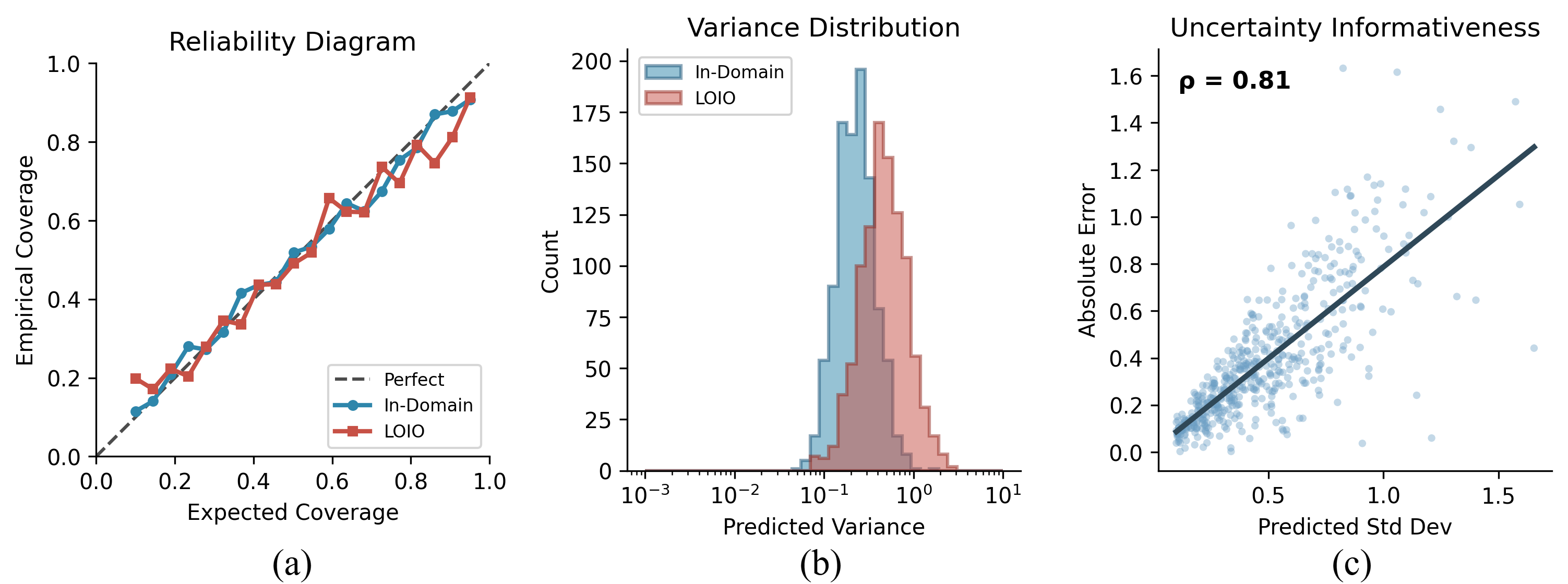}
    \captionof{figure}{Uncertainty calibration of SpikeSSL. Panel (a) shows reliability curves. Panel (b) compares variance distributions. Panel (c) plots std dev against absolute error.}
    \label{fig:uncertainty}
\end{minipage}
\end{figure*}

\begin{figure*}[t]
    \centering
  \begin{minipage}[t]{0.25\textwidth}
    \vspace{0pt}
    \centering
    \setlength{\tabcolsep}{2pt}
    \captionof{table}{Param-matched control. Scaling fails to close the gap.}
    \label{tab:params}
    \tiny
    \renewcommand\arraystretch{1.05}
    \begin{tabular}{@{}l c c@{}}
        \toprule
        Method & Mean Corr ($\uparrow$) & \#Params \\
        \midrule
        CASCADE-L          & 0.554 & 0.4M \\
        CASCADE-H          & 0.533 & 4.6M \\
        ENS\textsuperscript{2}-L & 0.576 & 2.1M \\
        ENS\textsuperscript{2}-H & 0.564 & 4.2M \\
        SpikeSSL-L & 0.649   & 7.2M \\
        \midrule
        \textbf{SpikeSSL}  & \textbf{0.654} & 3.9M \\
        \bottomrule
    \end{tabular}
  \end{minipage}
  \hfill
  \begin{minipage}[t]{0.722\textwidth}
    \vspace{0pt}
    \centering
    \includegraphics[width=\textwidth]{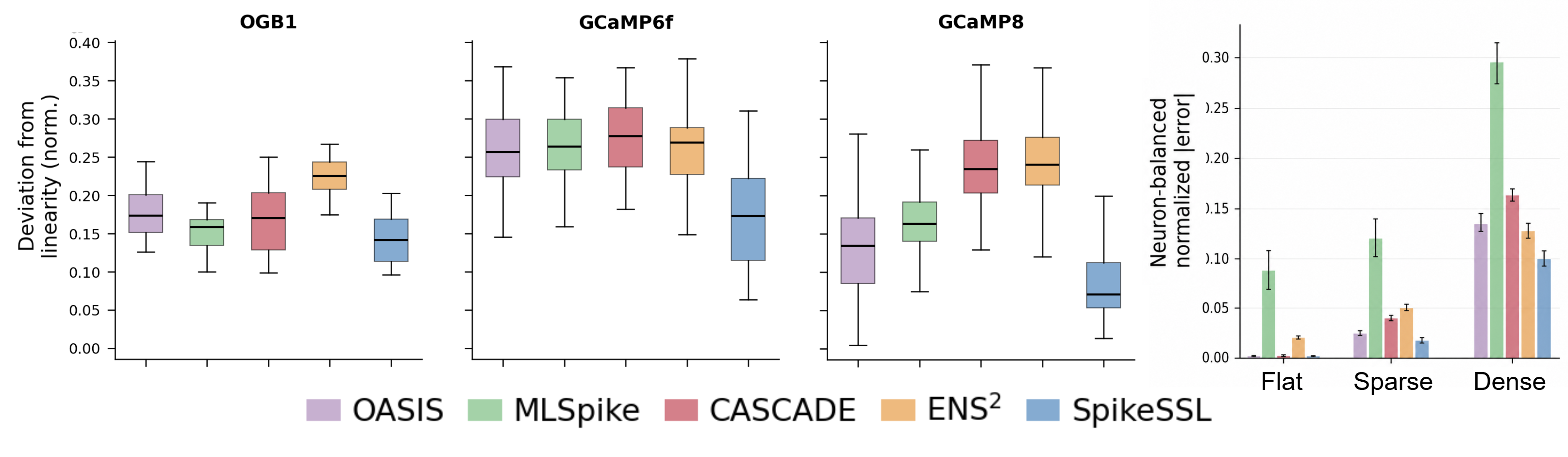}
      \caption{Linearity and activity-regime analysis. First three: deviation from linearity; Last: normalized error in flat, sparse, and dense windows.}
    \label{fig:linearity}
    \end{minipage}
\end{figure*}

\subsection{Ablation Studies}

We perform ablations under the LOIO protocol and report mean correlation across all five held-out splits in Table~\ref{tab:ablation}. V1 replaces AdaLN with a concatenation-based conditioning scheme, and V2 removes the signal-statistics branch from the conditioning vector. Together they confirm that both conditioning mechanisms contribute to cross-indicator adaptation. V3 replaces the bidirectional IIR module with a standard temporal convolutional block, supporting the value of the dynamics-informed multi-scale backbone. V4 removes the heteroscedastic variance head and $\beta$-NLL reweighting, while V5 removes the binary gate head and its cross-entropy loss, confirming that both heads improve predictive quality beyond their primary roles in uncertainty calibration and spike denoising.

\subsection{Discussion}
\textbf{Uncertainty Calibration}\quad SpikeSSL's heteroscedastic variance head provides calibrated per-frame uncertainty in addition to point estimates. Figure~\ref{fig:uncertainty} shows that confidence intervals remain close to empirical coverage, that predicted variance increases under LOIO shift, and that higher predicted standard deviation is associated with larger absolute error. Taken together, these observations suggest that the variance head captures meaningful epistemic uncertainty rather than acting as a generic smoothing term. At the same time, this analysis is limited to the five-split LOIO benchmark considered here, so whether calibration remains equally reliable under more extreme acquisition shifts, such as substantially different noise statistics or sampling regimes, remains an open question.

\textbf{Architectural Gains Beyond Parameter Scaling}\quad Table~\ref{tab:params} is intended to test whether SpikeSSL's gains can be explained simply by model size. We scale CASCADE and ENS\textsuperscript{2} to larger parameter budgets and also evaluate a larger SpikeSSL variant. Increasing the baseline parameter counts indeed improves their performance, but they still do not reach SpikeSSL. In contrast, further scaling SpikeSSL does not yield additional gains, suggesting that its advantage comes from the dynamics-informed IIR backbone, indicator-aware conditioning, and uncertainty-aware objective rather than from simply stacking more parameters.

\textbf{Linearity and Response Scale of Spike Rate Estimation}\quad A useful spike inference model should keep predicted firing rates close to a linear, correctly scaled transform of the ground truth. Figure~\ref{fig:linearity}(a) summarizes this property on OGB-1, GCaMP6f, and GCaMP8, where SpikeSSL shows the smallest deviation from linearity across distinct kinetic regimes. Figure~\ref{fig:linearity}(b) further stratifies local 1-s windows into flat, sparse, and dense activity regimes, showing lower error for SpikeSSL during both silent and active periods. These results are consistent with the multi-scale IIR backbone, whose fast and slow channels help preserve response scale across indicator kinetics.

\section{Conclusion}
We presented SpikeSSL, a universal spike inference framework that narrows the cross-indicator generalization gap through two complementary mechanisms. A bidirectional IIR backbone with a multi-scale decay basis spanning the full range of physiologically relevant indicator time constants, combined with multi-modal conditioning, provides a structured inductive bias for the inverse problem. Simulation-augmented training further bridges gaps in the kinetic parameter space of unseen indicators. On five fixed evaluation splits built from public ground-truth datasets, SpikeSSL achieves state-of-the-art performance in both in-domain and zero-shot settings without target-domain ground truth. By providing a single model that generalizes across indicator types without retraining, SpikeSSL can reduce the experimental burden of ground-truth data collection and facilitate the adoption of new calcium indicators in neuroscience laboratories.

%% file: original_appendix_content.tex
\section{Supplementary Material}
This appendix consolidates the technical details and extended validation in one continuous supplement. Appendix~\ref{app:supp_arch} expands the conditioning encoder, dilated refinement module, Temporal Mixer, and SwiGLU FFN. Appendix~\ref{app:benchmark_tables} documents the recording sets and fixed evaluation splits, Appendix~\ref{app:sim_details} specifies the biophysical simulation pipeline, and Appendix~\ref{app:benchmark_details} details benchmark construction. The remaining subsections provide additional analyses of conditioning, domain breadth, data scaling, ablations, adaptation regimes, learned dynamics, simulation, and robustness.

\subsection{Supplementary architecture modules}\label{app:supp_arch}
This subsection expands three modules that are named in the main text but not visualized in Figure~\ref{fig:main}, namely the conditioning encoder, the dilated refinement module, and the Temporal Mixer and SwiGLU FFN sub-blocks within each SpikeSSL block.

\begin{figure}[h]
    \centering
    \includegraphics[width=1\linewidth]{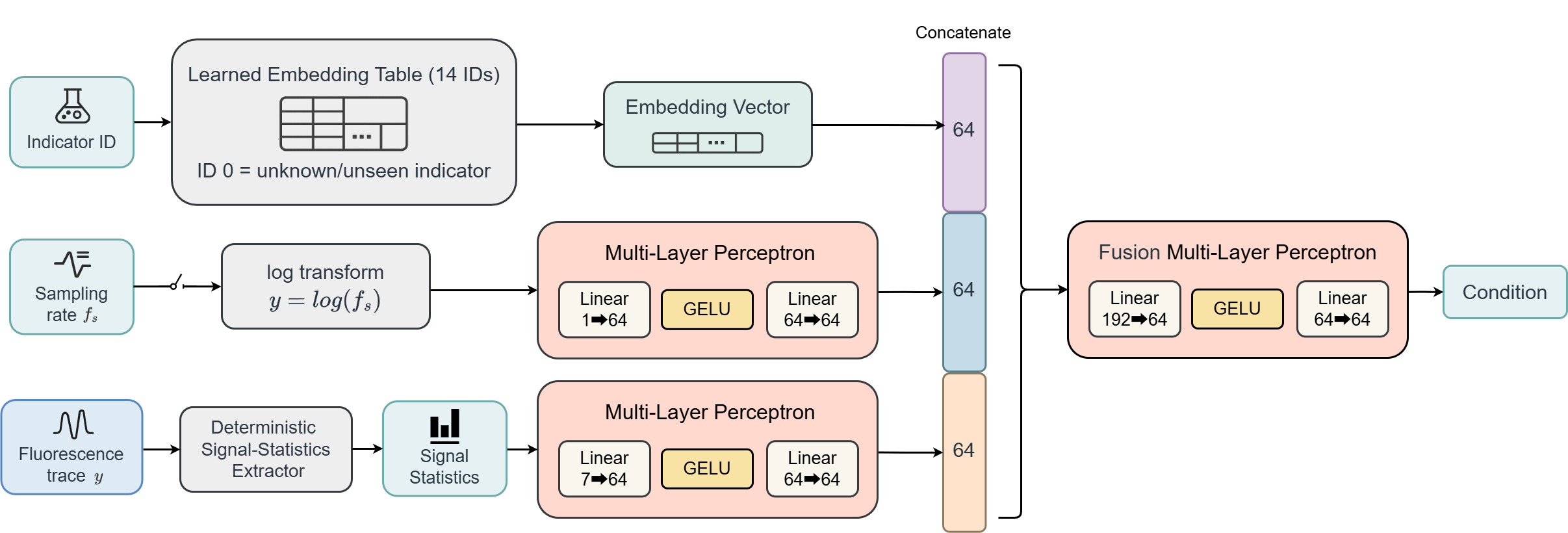}
    \caption{Conditioning encoder architecture. Three modality-specific branches independently encode their inputs before fusing them into a common representation. The indicator embedding branch maps discrete indicator identity to a 64-dimensional learned vector, where ID~0 is reserved for unknown or unseen indicators. The sampling-rate branch encodes $\log f_s$ through a two-layer MLP. The signal-statistics branch encodes the 7-dimensional descriptor comprising standard deviation, skewness, excess kurtosis, and autocorrelation at lags 1, 3, 5, and 10 through a two-layer MLP. The three 64-dimensional outputs are concatenated and fused by a $192 \to 64 \to 64$ MLP to produce the global conditioning vector $c \in \mathbb{R}^{64}$.}
    \label{fig:cond_enc}
\end{figure}

\subsubsection{Conditioning encoder internals}\label{app:cond_enc}
The conditioning encoder maps three modality-specific inputs through dedicated branches into the global conditioning vector $c \in \mathbb{R}^{64}$. The indicator branch uses a learned embedding table over 14 IDs, where ID~0 is reserved for unknown or unseen indicators and the remaining IDs cover named indicators including OGB-1, GCaMP6f, GCaMP6s, GCaMP8 variants, R-CaMP, jRCaMP1a, jGECO1a, and XCaMPgf, each projected to 64 dimensions by a linear layer. The sampling-rate branch maps $\log f_s$ through a two-layer MLP of hidden width 64, falling back to $\log 30$ when the frame rate is unavailable. The signal-statistics branch maps a 7-dimensional descriptor comprising standard deviation, skewness, excess kurtosis, and autocorrelation at lags 1, 3, 5, and 10 through a separate two-layer MLP of width 64. The three 64-dimensional branch outputs are concatenated and fused by a final $192 \to 64 \to 64$ MLP to produce $c$, keeping each modality isolated until the fusion stage so that indicator-specific metadata and data-driven statistics contribute through clearly separated pathways.

\subsubsection{Dilated refinement module}\label{app:dilated}
After the stack of conditioned SpikeSSL blocks and a final LayerNorm, the model applies a residual dilated convolution stack. With the default setting of three refinement layers, the dilation schedule is 1, 2, and 4, where each layer consists of a 1D convolution with kernel size 3 and matching dilation-dependent padding, followed by batch normalization, GELU, dropout, and a residual addition, expanding the temporal receptive field without downsampling. Whereas the bidirectional IIR scan captures long-range indicator-specific decay, the refinement stack provides complementary short-to-medium-range local correction capacity at full temporal resolution, functioning as a lightweight post-backbone smoother and sharpener.

\begin{figure}[h]
    \centering
    \includegraphics[width=\linewidth]{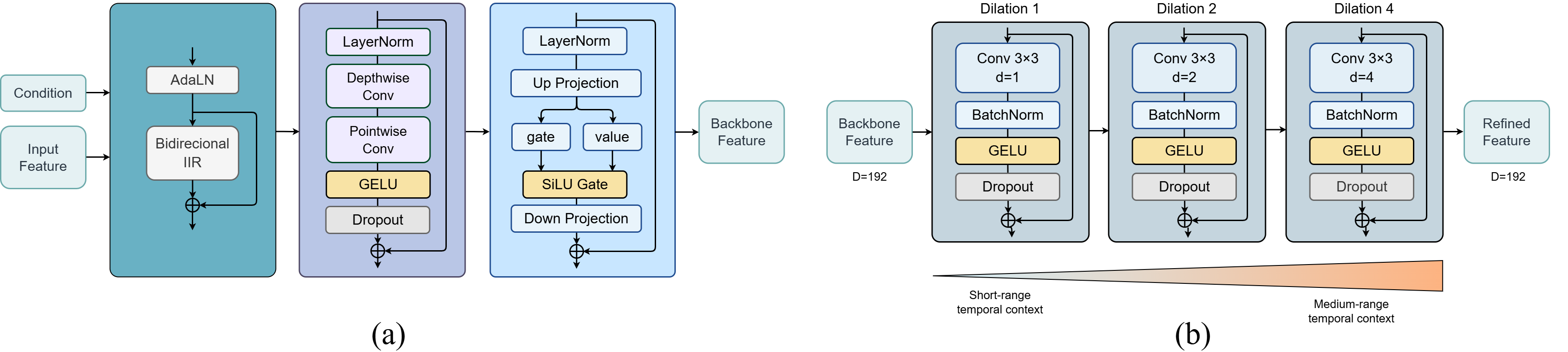}
    \caption{\textbf{(a)} SpikeSSL block sub-modules. The Temporal Mixer applies a depthwise 1D convolution (kernel width~7) followed by a pointwise convolution, GELU activation, and a residual connection, providing local temporal mixing without indicator-conditioned normalization. The SwiGLU FFN projects to twice the channel width, splits into gate and value tensors, computes $\mathrm{SiLU}(\mathrm{gate}) \odot \mathrm{value}$, and projects back to model dimension~$D$, followed by a residual connection. \textbf{(b)} Dilated refinement module comprising three residual layers with dilation factors 1, 2, and 4 that successively expand the temporal receptive field at full resolution, each containing a kernel-3 1D convolution, batch normalization, GELU, and dropout.}
    \label{fig:supp_blocks}
\end{figure}

\subsubsection{Temporal Mixer and SwiGLU FFN inside each block}\label{app:block_submods}
Each SpikeSSL block contains three sequential sub-modules. The first is the AdaLN-conditioned bidirectional IIR layer described in the main text. The second and third are deliberately left unconditioned. The Temporal Mixer applies LayerNorm to the block state, followed by a depthwise 1D convolution of kernel width 7, a pointwise 1D convolution, GELU activation, dropout, and a residual addition, providing local temporal mixing without an additional indicator-conditioned normalization path.

The SwiGLU FFN applies LayerNorm and projects the state to twice the expanded width, splits the projection into gate and value tensors, computes $\mathrm{SiLU}(\mathrm{gate}) \odot \mathrm{value}$, projects back to model dimension $D$, and adds the result residually. With the default configuration $D{=}192$ and expansion factor 2, the intermediate width is 384. AdaLN modulation is confined to the IIR branch, where indicator-specific decay kinetics have the clearest physical interpretation, while the Temporal Mixer and SwiGLU FFN serve as standard local and channel mixers.

\subsection{Benchmark data composition and split definitions}\label{app:benchmark_tables}
Table~\ref{tab:appendix_calcium_datasets} provides an overview of all benchmark datasets. Specifically, Table~\ref{tab:appendix_dataset_comp} categorizes the ground-truth recording sets by indicator family. Furthermore, Table~\ref{tab:appendix_families} lists the five fixed evaluation splits.

The recording-level inventory makes the source of benchmark heterogeneity explicit: the corpus combines chemical dyes and genetically encoded indicators across species, anatomical regions, induction methods, and sampling rates rather than treating an indicator as a single acquisition setup. Frame rates span approximately 2.5--500~Hz, and the recordings cover cortical, hippocampal, olfactory, and spinal preparations collected by multiple laboratories. Consequently, the benchmark tests variation in acquisition and biological context in addition to nominal sensor identity. Recording-level separation also prevents overlapping windows from the same acquisition from being divided between training and evaluation.

\begin{table}[H]
  \centering
  \caption{Overview of all ground truth recording sets.}
  \label{tab:appendix_calcium_datasets}
  \scriptsize
  \setlength{\tabcolsep}{5pt}
  \renewcommand\arraystretch{1.08}
  \begin{adjustbox}{min width=0.78\linewidth,max width=\linewidth,center}
\begin{tabular}
  {@{} p{0.03\linewidth} p{0.1\linewidth} p{0.21\linewidth} p{0.08\linewidth} p{0.13\linewidth} p{0.08\linewidth} p{0.03\linewidth} l@{}}
    \toprule
    Set & Calcium \newline indicator & Induction \newline method & Animal \newline model & Anatomical \newline region & Frame rate \newline (Hz) & $N$ & Source \newline paper \\
    \midrule
    1  & OGB-1 & Acute injection & Mouse & V1 & 11.3 & 11 & Theis et al., 2016 \\
    2  & OGB-1 & Injection + tg(tdTomato-CaMKII$\alpha$) & Mouse & V1 & 15.6 & 16 & Kwan and Dan, 2012 \\
    3  & Cal-520 & Acute injection & Mouse & S1 & 500.0 & 8 & Tada et al., 2014 \\
    4  & OGB-1 & Acute injection & Zebrafish & pDp & 7.7 & 15 & Rupprecht et al., 2021 \\
    5  & Cal-520 & Acute injection & Zebrafish & pDp & 7.8 & 5 & Rupprecht et al., 2021 \\
    6  & GCaMP6f & tg(NeuroD) & Zebrafish & aDp & 30.0 & 8 & Rupprecht et al., 2021 \\
    7  & GCaMP6f & tg(NeuroD) & Zebrafish & dD & 30.0 & 10 & Rupprecht et al., 2021 \\
    8  & GCaMP6f & tg(NeuroD) & Zebrafish & OB & 30.0 & 9 & Rupprecht et al., 2021 \\
    9  & GCaMP6f & AAV & Mouse & V1 & 60.1 & 11 & Chen et al., 2013 \\
    10 & GCaMP6f & tg(Emx1) & Mouse & V1 & 160.1 & 23 & Huang et al., 2019 \\
    11 & GCaMP6f & tg(Cux2) & Mouse & V1 & 158.3 & 25 & Huang et al., 2019 \\
    12 & GCaMP6s & tg(tetOs) & Mouse & V1 & 151.6 & 6 & Huang et al., 2019 \\
    13 & GCaMP6s & tg(Emx1) & Mouse & V1 & 157.5 & 26 & Huang et al., 2019 \\
    14 & GCaMP6s & AAV & Mouse & V1 & 60.1 & 7 & Chen et al., 2013 \\
    15 & GCaMP6s & AAV & Mouse & V1 & 59.1 & 9 & Theis et al., 2016 \\
    16 & GCaMP6s & AAV & Mouse & V1 & 59.1 & 9 & Theis et al., 2016 \\
    17 & GCaMP5k & AAV & Mouse & V1 & 50.0 & 9 & Akerboom et al., 2012 \\
    18 & R-CaMP1.07 & tg(Grik4-cre) + AAV & Mouse & CA3 & 20.0 & 4 & Schoenfeld et al., 2021 \\
    19 & R-CaMP1.07 & AAV & Mouse & S1 & 15.0 & 9 & Bethge et al., 2017 \\
    20 & jRCaMP1a & AAV & Mouse & V1 & 15.0 & 10 & Dana et al., 2016 \\
    21 & jRGECO1a & AAV & Mouse & V1 & 29.8 & 11 & Dana et al., 2016 \\
    22 & OGB-1 & Injection + tg(GFP-GIN) & Mouse & V1 (SST) & 15.6 & 5 & Kwan and Dan, 2012 \\
    23 & OGB-1 & Injection + tg(tdTomato-PV) & Mouse & V1 (PV) & 15.6 & 7 & Kwan and Dan, 2012 \\
    24 & GCaMP6f & tg(PV-cre) + AAV & Mouse \newline (in vitro) & V1 (PV) & 26.6 & 13 & Khan et al., 2018 \\
    25 & GCaMP6f & tg(SOM-cre) + AAV & Mouse \newline (in vitro) & V1 (SST) & 26.6 & 17 & Khan et al., 2018 \\
    26 & GCaMP6f & tg(VIP-cre) + AAV & Mouse \newline (in vitro) & V1 (VIP) & 26.6 & 11 & Khan et al., 2018 \\
    27 & GCaMP6f & tg(tdTomato-PV) + AAV & Mouse & V1 (PV) & 30.0 & 4 & Rupprecht et al., 2021 \\
    28 & XCaMPgf & AAV & Mouse & V1 & 122 & 8 & Zhang et al., 2023 \\
    29 & GCaMP7f & AAV & Mouse & V1 & 122 & 22 & Zhang et al., 2023 \\
    30 & GCaMP8f & AAV & Mouse & V1 & 122 & 36 & Zhang et al., 2023 \\
    31 & GCaMP8m & AAV & Mouse & V1 & 122 & 42 & Zhang et al., 2023 \\
    32 & GCaMP8s & AAV & Mouse & V1 & 122 & 39 & Zhang et al., 2023 \\
    33 & Mixed (GCaMP7f /8/ XCaMPgf) & AAV & Mouse & V1 (Interneurons) & 122 & 17 & Zhang et al., 2023 \\
    40 & GCaMP6s & tg(VGlut2-Cre) & Mouse & Spinal cord & $\sim$2.5 & 21 & Rupprecht et al., 2025 \\
    41 & GCaMP6s & tg(Viaat-Cre) & Mouse & Spinal cord & $\sim$2.5 & 23 & Rupprecht et al., 2025 \\
    \bottomrule
  \end{tabular}
\end{adjustbox}
\end{table}

\begin{table}[H]
  \centering
  \caption{Ground-truth recording sets used in the benchmark.}
  \label{tab:appendix_dataset_comp}
  \tiny
  \setlength{\tabcolsep}{5pt}
  \renewcommand\arraystretch{0.95}
  \begin{adjustbox}{min width=0.78\linewidth,max width=\linewidth,center}
\begin{tabular}{@{}p{0.25\linewidth}p{0.17\linewidth}p{0.38\linewidth}r@{}}
    \toprule
    Category & Indicator & Recording sets & $N$ \\
    \midrule
    Chemical dye & OGB-1 & DS01, DS02, DS04, DS22, DS23 & 64 \\
    Chemical dye & Cal-520 & DS03, DS05 & 13 \\
    Legacy GECI & GCaMP5k & DS17 & 9 \\
    Fast GCaMP6 & GCaMP6f & DS06--DS11, DS24--DS27 & 124 \\
    Slow GCaMP6 & GCaMP6s & DS12--DS16 & 57 \\
    GCaMP7 & GCaMP7f & DS29 & 22 \\
    GCaMP8 variants & GCaMP8f, GCaMP8m, GCaMP8s & DS30, DS31, DS32 & 117 \\
    Red-shifted GECI & R-CaMP & DS18, DS19 & 13 \\
    Red-shifted GECI & jRCaMP1a & DS20 & 9 \\
    Red-shifted GECI & jGECO1a & DS21 & 11 \\
    Other GECI & XCaMPgf & DS28 & 8 \\
    Interneurons & GCaMP7f, GCaMP8f, GCaMP8m, GCaMP8s, XCaMPgf & DS33 & 17 \\
    Spinal cord (excit.) & GCaMP6s & DS40 & 21 \\
    Spinal cord (inhib.) & GCaMP6s & DS41 & 23 \\
    \bottomrule
  \end{tabular}
\end{adjustbox}
\end{table}

Grouping these recordings by indicator family shows how the 33 source datasets contribute complementary kinetic regimes. The fixed splits below then isolate five representative transfer settings while preserving recording-level separation.

\begin{table}[H]
  \centering
  \caption{Fixed evaluation splits. Source-side recordings are excluded from training under LOIO; evaluation recordings are held out in all settings.}
  \label{tab:appendix_families}
  \scriptsize
  \setlength{\tabcolsep}{5pt}
  \renewcommand\arraystretch{1.08}
  \begin{adjustbox}{min width=0.78\linewidth,max width=\linewidth,center}
\begin{tabular}{@{}p{0.20\linewidth}p{0.26\linewidth}p{0.22\linewidth}p{0.22\linewidth}@{}}
    \toprule
    Split & Held-out under LOIO & Eval.\ recordings & Contrast \\
    \midrule
    OGB-1 & DS01, DS22, DS23 & DS02, DS04 & Chemical-dye transfer \\
    GCaMP6s & DS12, DS13, X-DS11, X-DS12 & DS14, DS15, DS16 & Slow-decay GECI \\
    GCaMP8 variants & DS30, DS31 & DS32 & GCaMP8f/8m$\to$8s \\
    Red-shifted & DS18, DS19, DS20 & DS21 & Cross-indicator transfer \\
    Spinal cord & DS41 & DS40 & Brain-to-spinal-cord \\
    \bottomrule
  \end{tabular}
\end{adjustbox}
\end{table}

\subsection{Loss function ablation under \texttt{+sim} LOIO}\label{app:loss_ablation}
Table~\ref{tab:loss_ablation} compares four loss configurations on the five \texttt{+sim} LOIO splits. Adding each auxiliary term consistently improves mean correlation, and the full three-component setup achieves the highest score on every split.

\begin{table}[H]
  \centering
  \caption{Mean Pearson correlation under \texttt{+sim} LOIO for SpikeSSL trained with progressive loss additions. All runs use identical data and hyperparameters, and only the active loss terms differ. \textbf{Bold}: best per column.}
  \label{tab:loss_ablation}
  \scriptsize
  \setlength{\tabcolsep}{6pt}
  \renewcommand\arraystretch{1.1}
  \begin{adjustbox}{min width=0.78\linewidth,max width=\linewidth,center}
\begin{tabular}{@{}lcccccc@{}}
    \toprule
    Loss configuration & V1-OGB & V1-GCaMP6s & V1-GCaMP8 & Other-RCaMP & SC-GCaMP6s & Mean \\
    \midrule
    Huber regression only & 0.527 & 0.445 & 0.861 & 0.703 & 0.622 & 0.632 \\
    + variance head ($\beta$-NLL) & 0.536 & 0.461 & 0.872 & 0.712 & 0.641 & 0.644 \\
    + gate BCE & 0.540 & 0.467 & 0.876 & 0.717 & 0.651 & 0.650 \\
    Full (+ mean-rate) & \textbf{0.547} & \textbf{0.476} & \textbf{0.886} & \textbf{0.725} & \textbf{0.668} & \textbf{0.660} \\
    \bottomrule
  \end{tabular}
\end{adjustbox}
\end{table}

\subsection{Biophysical simulation pipeline specification}\label{app:sim_details}
\subsubsection{Spike Time Simulation}

To construct realistic spike timings, we implemented two distinct strategies, namely drawing inter-spike intervals from predefined probability distributions and simulating membrane potential dynamics via the Izhikevich neuron model.

In the first strategy, the total spike count for each trial is determined by a target mean firing rate $\lambda_\mathrm{spike}$ and a randomly generated recording duration $T$. Three stochastic point-process models cover distinct firing regimes. For standard stochastic firing, the spike count follows a Poisson distribution $N \sim \mathrm{Poisson}(\lambda_\mathrm{spike} T)$ and inter-spike intervals (ISIs) are drawn independently. A Gamma distribution $\Delta t_k \sim \Gamma(\kappa, \theta)$ generalizes this to model the relative refractory period, reducing to the standard exponential when $\kappa = 1$ and penalizing short ISIs when $\kappa > 1$. For quasi-periodic firing, a linear interpolation introduces stochastic perturbations into a uniform spike train, $\Delta t_k = (1-r) \cdot T_\mathrm{period} + r \cdot T_\mathrm{rand}$, where $T_\mathrm{period} = 1/\lambda$ and $r \in [0,1]$ controls the degree of randomness. For bursty firing, burst onsets follow a Poisson process with rate $\lambda_\mathrm{burst}$, the spike count per burst follows $n_i \sim \mathrm{Poisson}(\mu_\mathrm{burst})$, and intra-burst ISIs are generated by stochastic jitter around a sub-millisecond mean interval.

In the second strategy, we numerically integrate the Izhikevich neuron model over a duration $T$ to efficiently generate biophysically realistic spike trains. By capturing the rapid depolarization and slow recovery of the membrane potential, this model faithfully reproduces complex firing dynamics. To reflect physiological stochasticity, the driving current $I(t)$ is formulated as a multi-scale composite synaptic drive:
$$I(t) = I_{\text{base}} + I_{\text{slow}}(t) + I_{\text{med}}(t) + I_{\text{fast}}(t),$$
where $I_{\text{base}}$ denotes the baseline current profile, while the $slow$, $med$, and $fast$ terms represent the respective frequency components of the synaptic perturbations.The temporal evolution of the fast membrane potential $v$ and the slow recovery variable $u$ under this driving current is governed by the following system of ordinary differential equations:
\begin{align*}
\frac{dv}{dt} &= 0.04v^2 + 5v + 140 - u + I,\\
\frac{du}{dt} &= a(bv - u).
\end{align*}
Upon reaching the spike apex $(v\ge 30mV)$, the state variables are instantaneously reset:  $v \gets c$ and $u \gets u + d$. The dimensionless parameter set $(a, b, c, d)$ governs these intrinsic dynamics, enabling the simulation of distinct neuronal subclasses, including Regular Spiking (RS), Intrinsically Bursting (IB), Chattering (CH), and Fast Spiking (FS) neurons.

\subsubsection{Fluorescence Signal Simulation}

In the MLSpike framework, the normalized intracellular calcium concentration $c(t)$ is modeled as a first-order linear system driven by the spike train $s(t) = \sum_{k} \delta(t - t_k)$, which is a set of Dirac functions placed at spike times $t_k$. The calcium level decays exponentially with a time constant $\tau$, governed by the differential equation:
$$\frac{dc}{dt} = -\frac{1}{\tau}c + s(t).$$

However, the assumption of a constant unitary response deviates from actual biological processes and contradicts experimentally observed data. To address this, we define the input $s(t)$ as a sequence of weighted discrete events, where the calcium influx elicited by the $k$-th spike at time $t_k$ is scaled by a dynamic weight $w(t_k)$. Consequently, the dynamics of $c(t)$ are modified to:
$$\frac{dc}{dt} = -\frac{1}{\tau}c + s(t)*w(t),$$
where the dynamic weight is formulated as:
$$w(t) = R(t) \cdot \max(0, 1 + \epsilon_k).$$
Here, $\epsilon_k \sim \mathcal{N}(0, \sigma_A^2)$ represents the stochastic amplitude jitter drawn from a normal distribution, reflecting the natural variability of calcium influx per spike. The variable $R(t) \in [0, 1]$ serves as a depletion variable representing the fraction of the available calcium response pool. When an action potential occurs at $t = t_k$, the available pool is instantaneously depleted by a constant proportion $d$ (the depletion factor):
$$R(t_k^+)= R(t_k) \cdot (1 - d),$$
where $R(t_k^+)$ denotes the instantaneous value of $R(t)$ immediately after the spike occurs at time $t_k$. During the inter-spike intervals ($t \neq t_k$), the temporal evolution is a continuous exponential recovery towards its maximum capacity of $1$:
$$\frac{dR}{dt} = \frac{1}{\tau_{rec}} (1 - R(t)),$$
where $\tau_{rec}$ is the recovery time constant.

Subsequent to the influx of free calcium, the observable fluorescence signal is fundamentally determined by the concentration of calcium-bound indicator molecules. The binding between calcium ions and indicator molecules $[\mathrm{B}]$ obeys mass-action kinetics, as illustrated by the reaction:
$$[\mathrm{Ca}]^n + [\mathrm{B}]
\overset{k_{\mathrm{on}}}{\underset{k_{\mathrm{off}}}{\rightleftharpoons}}
[\mathrm{Ca}^n\mathrm{B}],$$
where $k_{\mathrm{on}}$ and $k_{\mathrm{off}}$ are the association and dissociation rates, respectively, and $n$ is the Hill coefficient representing the degree of cooperativity. This formulation can accommodate both genetically encoded calcium indicators (GECIs, e.g., GCaMP, where $n > 1$) and synthetic chemical dyes (e.g., OGB-1, where $n = 1$).
According to the law of mass action, the rate of change for the bound product concentration is given by:
$$\frac{d\left[Ca^nB\right]}{dt} = k_{\text{on}} \left[Ca\right]^{n} \left[B\right] - k_{\text{off}} \left[Ca^nB\right].$$
By defining the normalized bound indicator fraction $p(t)$ as:
$$p = \frac{1}{\gamma} \frac{ \left[ Ca^nB \right] - \left[ Ca^nB \right]_0 }{ \left[ B \right]_T - \left[ Ca^nB \right]_0 },$$
the temporal evolution of $p(t)$ under standard constant-rate assumptions is governed by the following parameterized differential equation:
$$\frac{dp}{dt}=\frac{1}{\tau_{on}}\left[1+\gamma((c_0+c)^n-c_0^n)\right]\left(\frac{(c_0+c)^n-c_0^n}{1+\gamma((c_0+c)^n-c_0^n)}-p\right),$$
where $\tau_{on}$ is the binding time constant at baseline, $c_0$ is the normalized resting calcium level, and $\gamma$ is the saturation parameter limiting the maximum bound fraction.

To extend this standard formulation and mechanistically characterize the conformational locking of the indicator molecules, we relax the constant-rate assumption for $k_{\mathrm{off}}$. Specifically, the dissociation rate is modeled as a dynamically decelerating variable that explicitly depends on the instantaneous occupancy fraction $p(t)$:
$$k_{\text{off}}(t) = k_{\text{off},0} \cdot \max\left(0.1, 1 - \alpha \cdot p(t)\right),$$
where $k_{\text{off},0}$ is the basal dissociation rate and $\alpha$ determines the deceleration magnitude.

Next, we apply a phenomenological polynomial mapping to the dynamically derived variable $p(t)$. This transformation yields the nonlinearly scaled bound fraction $p_{\text{final}}(t)$:
$$p_{\text{final}}(t) = 
\begin{cases} 
c_3 p(t)^3 + c_2 p(t)^2 + c_1 p(t), & \text{if } p(t) \le 1 \\[6pt]
1 + \kappa \cdot \left( p(t) - 1 \right), & \text{if } p(t) > 1
\end{cases}$$
where the constraint $c_1 = 1 - c_2 - c_3$ ensures gain conservation at $p = 1$, and  $\kappa = 3c_3+2c_2+c_1$ represents the tangent slope at $p = 1$, facilitating a smooth linear extrapolation that guarantees numerical stability outside the standard dynamic range.

Furthermore, to account for the slow continuous baseline drifts inherent to realistic scenarios, we model the baseline fluorescence $B(t)$ as a Wiener process:
$$\frac{dB(t)}{dt} = \eta \cdot \xi_B(t),$$
where $\eta$ scales the amplitude of the baseline drift and $\xi_B(t)$ denotes standard Gaussian white noise.

Ultimately, the observed fluorescence intensity $F(t)$ is formulated as the integrated effect of the drifting baseline, the dynamic bound indicator concentration, and the measurement noise:
$$F(t) = B(t) \left( 1 + A \cdot p(t) \right) + \sigma \cdot \xi_{F}(t),$$
where $A$ represents the maximum fluorescence transient amplitude, and  $\sigma \cdot \xi_{F}(t)$ accounts for the additive Gaussian noise with variance $\sigma^2$.

\subsection{Detailed benchmark construction}\label{app:benchmark_details}
The benchmark uses fixed recording-set partitions rather than random neuron-level resampling. In the in-domain setting, the source-side recordings listed in Table~\ref{tab:appendix_families} are included during training and the corresponding evaluation recordings are held out for testing. In the LOIO setting, the source-side recordings from the target evaluation family are removed from training, while the evaluation recordings remain unchanged. All other real source-domain recordings remain available for training, so the only controlled difference between LOIO runs is the held-out indicator or tissue family.

The five splits in Table~\ref{tab:appendix_families} are the evaluation units behind the in-domain and LOIO benchmark tables in the main paper. The held-out evaluation recordings are fixed at the recording-set level, which prevents different methods from being compared under different neuron-level samples.

Before training and testing, each fluorescence trace is windowed into 128-frame segments with 30 frames of left context, 30 frames of right context, and overlap size 2. We use raw fluorescence as the input representation, do not apply offline denoising, and do not resample traces before evaluation. When simulation augmentation is enabled, simulated traces are appended to the training pool only and are never included in any evaluation split.

\subsection{Additional results and analyses}

\subsubsection{Representative transfer functions}\label{app:transfer_function}

\begin{figure}[htbp]
    \centering
    \includegraphics[width=0.9\linewidth]{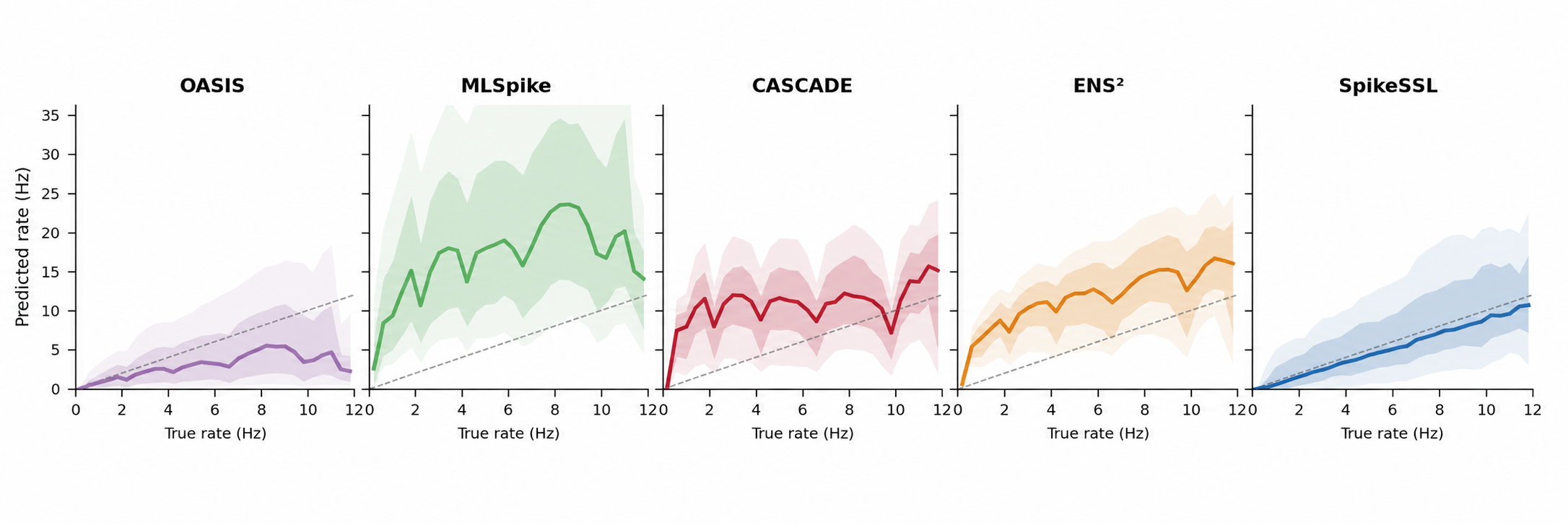}
    \caption{Representative transfer functions for five commonly used indicators. The shaded region denotes inter-neuron variability, and the dashed diagonal indicates the identity mapping. Indicators are selected because they span distinct kinetic regimes.}
    \label{fig:transfer_function}
\end{figure}

Figure~\ref{fig:transfer_function} indicates that the dominant failure mode is not a uniform offset but an indicator-dependent distortion of gain. For OASIS and MLSpike, the predicted rate increases too slowly once the true rate enters the moderate and high regime, which is consistent with the strong saturation visible in the curves. CASCADE and ENS\textsuperscript{2} recover more dynamic range, but their mappings still bend away from the identity and exhibit broader inter-neuron variability, indicating that rate calibration remains unstable across indicators. SpikeSSL stays closer to the identity over a wider operating range and maintains a narrower spread across neurons, suggesting that its advantage comes from preserving response gain more faithfully rather than from isolated corrections at a few firing-rate values.

\subsubsection{Long-trial LOIO visualizations}

To complement the per-window correlation scores reported in the main benchmark, we provide long-trial visualisations that span the full recording duration. We include one representative neuron from each of the six held-out calcium indicators in the LOIO evaluation, namely the chemical dye OGB-1, the early and improved fast green GECIs GCaMP5k and GCaMP7f, the slow green GECI GCaMP6s, the red-shifted GECI jGECO1a, and the next-generation fast green GECI GCaMP8f. For every neuron we display the recorded fluorescence trace together with spike-rate predictions from all six methods over the entire recording, with the smoothed ground-truth spike rate overlaid in grey for reference. The Pearson correlation coefficient $r$ reported in each panel is computed over the full trace. These visualisations corroborate the in-domain and LOIO benchmarks in the main paper, and additionally show that SpikeSSL with simulation augmentation tracks both isolated bursts and prolonged elevated firing across indicators with markedly different kinetics, while the remaining gap that simulation augmentation closes is most pronounced for indicators whose kinetics deviate substantially from the supervised training pool, such as jGECO1a and GCaMP5k.

\begin{figure}[htbp]
\centering
\includegraphics[width=0.9\linewidth]{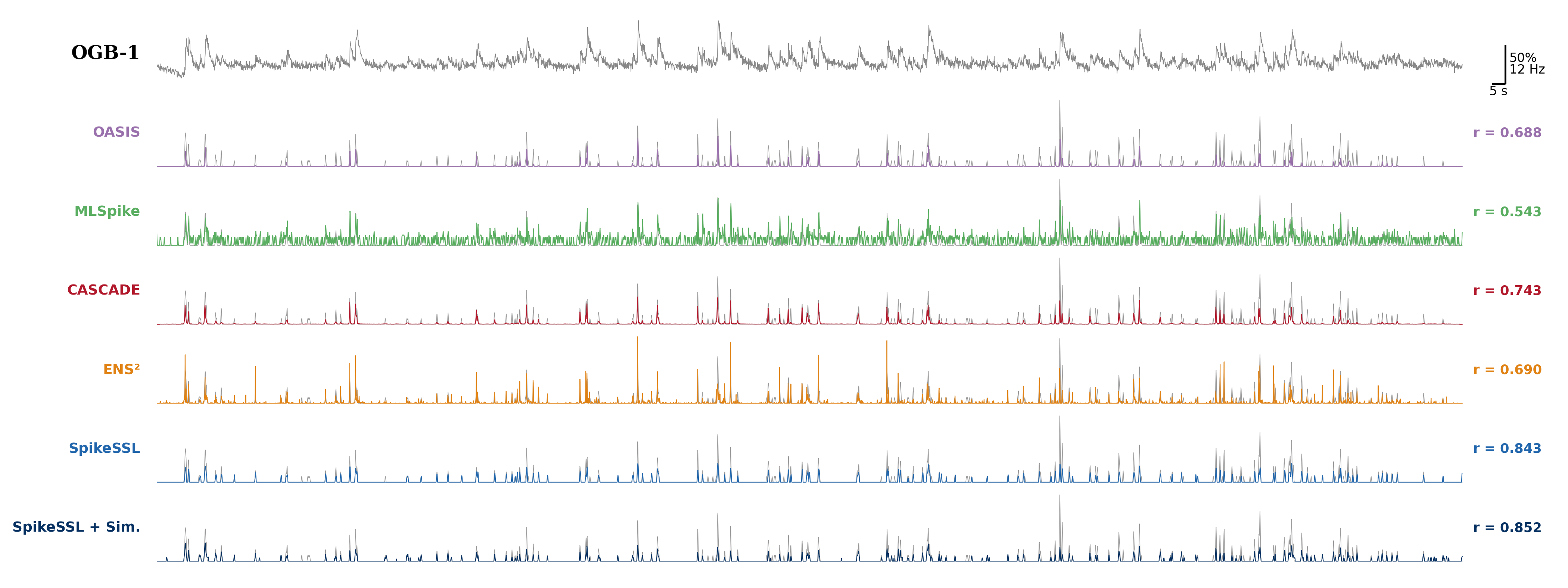}
\caption{
  LOIO spike inference on a representative OGB-1 neuron
  (chemical dye with slow calcium decay kinetics).
  The top row shows the recorded $\Delta F/F$ trace.
  Subsequent rows display spike-rate estimates produced by
  OASIS, MLSpike, CASCADE, ENS$^2$, \SpikeSSL, and \SpikeSSL{}+Sim.,
  each overlaid with the smoothed ground-truth spike rate (grey).
  The Pearson correlation coefficient $r$ is computed over the full recording duration.
}
\label{fig:loio_full_OGB1}
\end{figure}

\begin{figure}[htbp]
\centering
\includegraphics[width=0.9\linewidth]{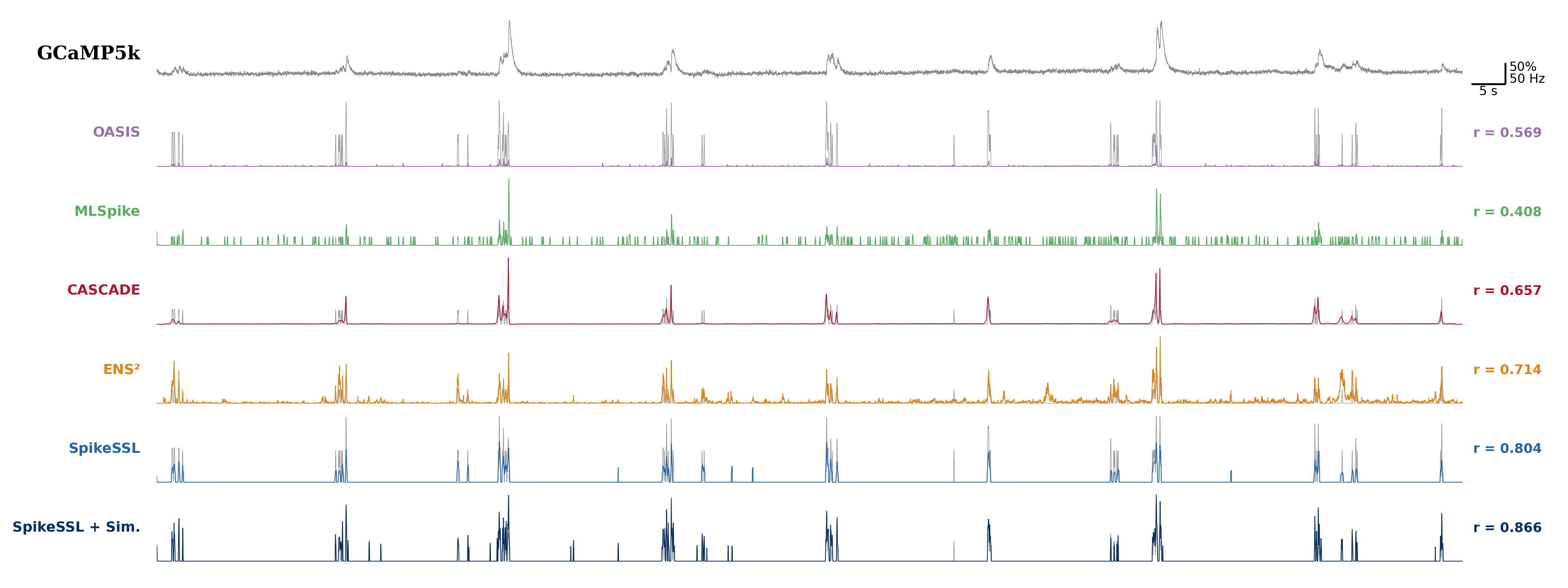}
\caption{
  LOIO spike inference on a representative GCaMP5k neuron
  (an early-generation fast green GECI).
  Panel layout follows Figure~\ref{fig:loio_full_OGB1}.
}
\label{fig:loio_full_GCaMP5k}
\end{figure}

\begin{figure}[htbp]
\centering
\includegraphics[width=0.9\linewidth]{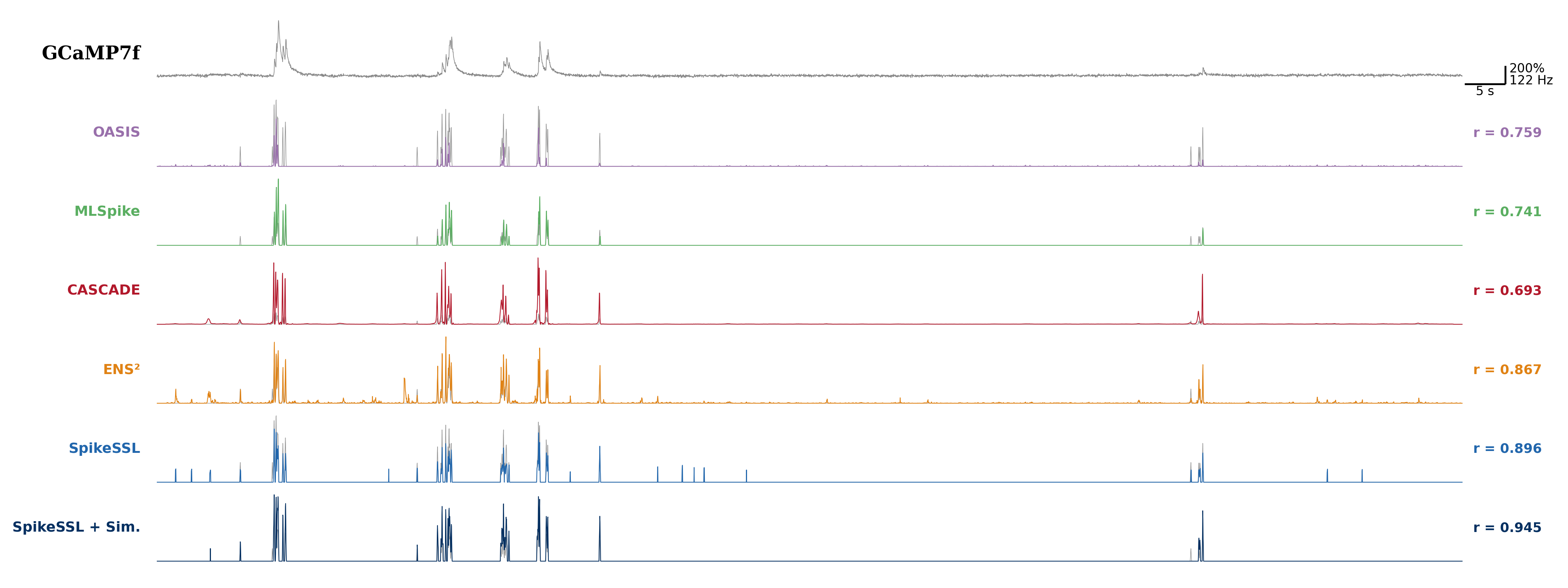}
\caption{
  LOIO spike inference on a representative GCaMP7f neuron
  (an improved fast green GECI with enhanced sensitivity).
  Panel layout follows Figure~\ref{fig:loio_full_OGB1}.
}
\label{fig:loio_full_GCaMP7f}
\end{figure}

\begin{figure}[htbp]
\centering
\includegraphics[width=0.9\linewidth]{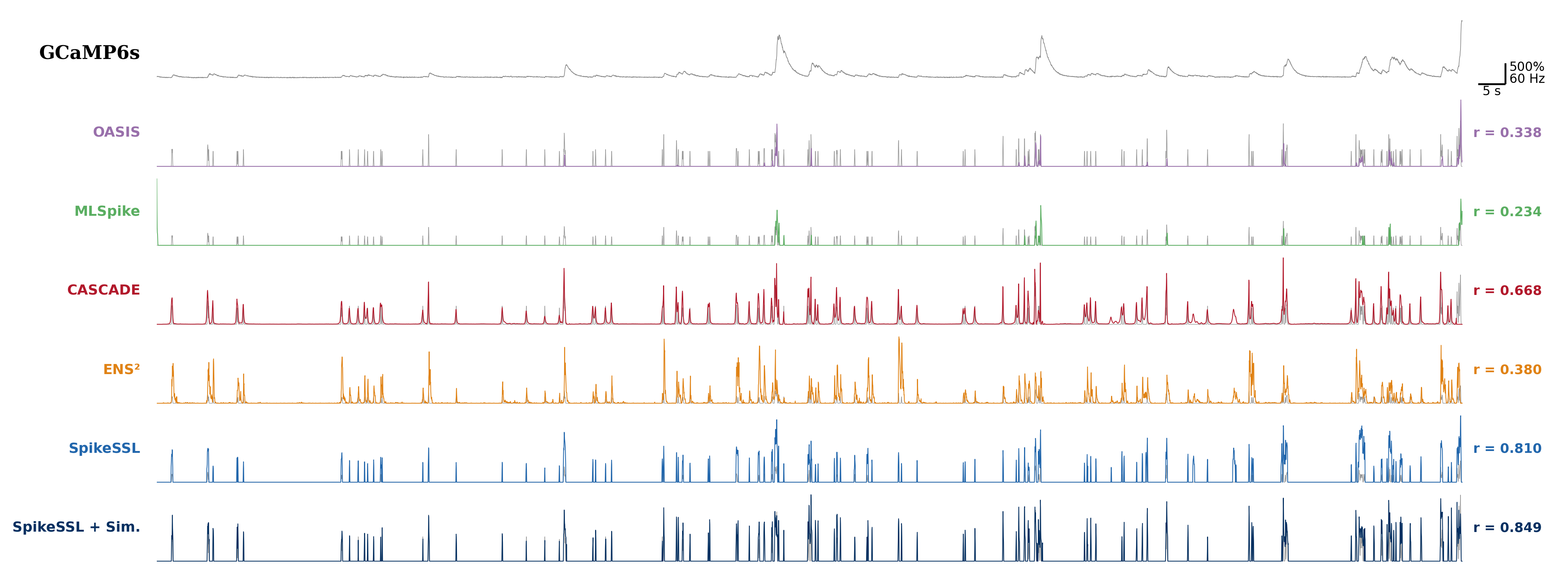}
\caption{
  LOIO spike inference on a representative GCaMP6s neuron
  (a slow GECI recorded in a low-activity regime).
  Panel layout follows Figure~\ref{fig:loio_full_OGB1}.
}
\label{fig:loio_full_GCaMP6s}
\end{figure}

\begin{figure}[htbp]
\centering
\includegraphics[width=0.9\linewidth]{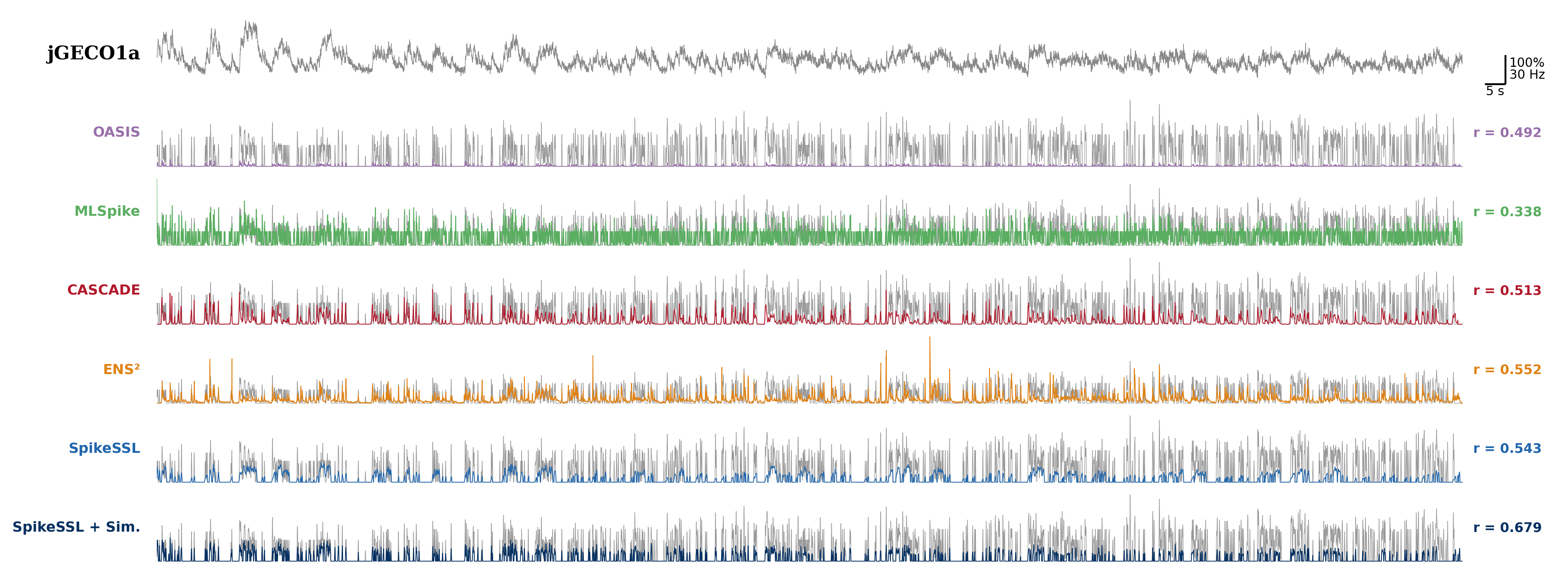}
\caption{
  LOIO spike inference on a representative jGECO1a neuron
  (a red-shifted GECI with distinct spectral and kinetic properties).
  Panel layout follows Figure~\ref{fig:loio_full_OGB1}.
}
\label{fig:loio_full_jGECO1a}
\end{figure}

\begin{figure}[htbp]
\centering
\includegraphics[width=0.9\linewidth]{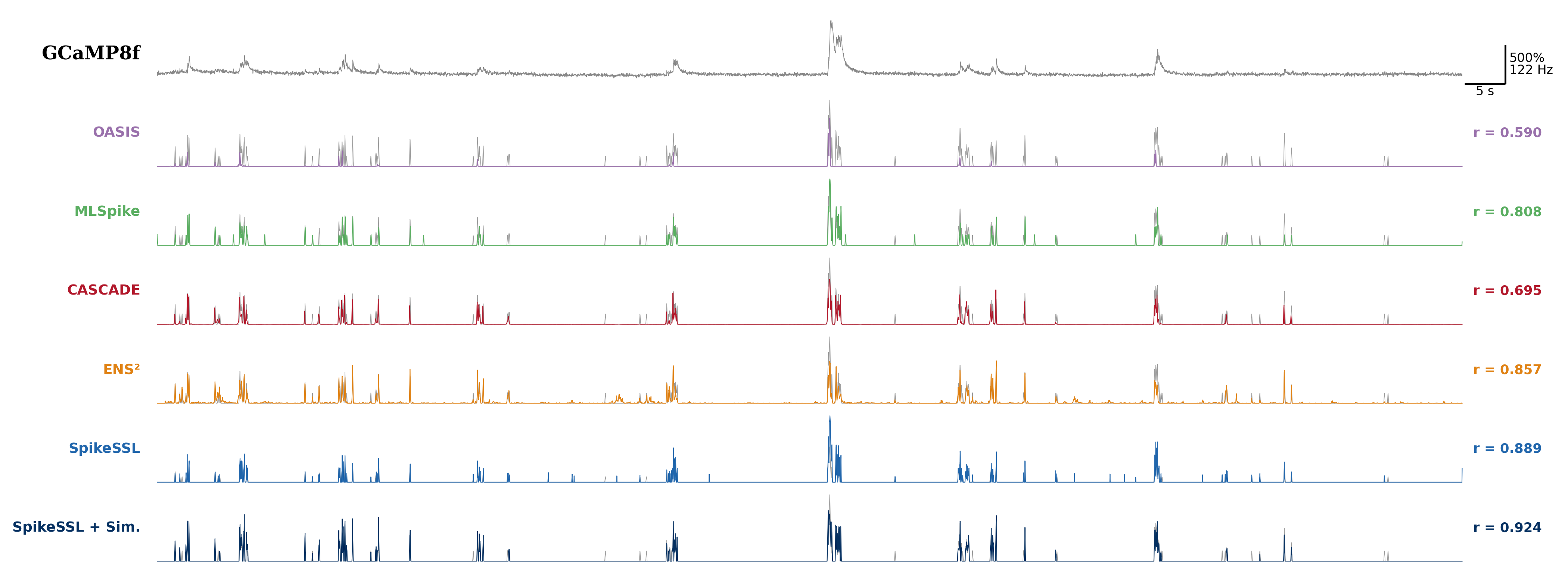}
\caption{
  LOIO spike inference on a representative GCaMP8f neuron
  (a next-generation fast GECI with high signal-to-noise ratio).
  Panel layout follows Figure~\ref{fig:loio_full_OGB1}.
}
\label{fig:loio_full_GCaMP8f}
\end{figure}

%% file: rebuttal_appendix_content.tex
The following subsections present additional controls, robustness studies, and
mechanistic analyses that complement the main results. All experiments
use the fixed recording-set partitions described in Appendix~\ref{app:benchmark_details}.
Unless stated otherwise, correlation denotes Pearson correlation of predicted
and ground-truth spike rates, higher is better, and error and absolute bias are
lower is better.

\subsection{Test-Time Conditioning Uses No Target-Domain Pool}
\label{app:conditioning_controls}

The seven trace statistics are computed independently from the current input
window. Thus, LOIO inference uses one unlabeled input window, no target-domain
pool, no target labels, and no parameter update. Pooling statistics across
multiple target windows does not improve performance (Table~\ref{tab:stat_pool}),
supporting that the method is per-window conditional inference rather than
few-shot adaptation.

\begin{table}[H]
\centering
\caption{Effect of pooling unlabeled target windows for trace statistics under LOIO.}
\label{tab:stat_pool}
\small
\begin{adjustbox}{min width=0.78\linewidth,max width=\linewidth,center}
\begin{tabular}{lcccc}
\toprule
& Per-window & Pool 5 & Pool 20 & Pool all \\
\midrule
Mean correlation & \textbf{0.654} & 0.649 & 0.650 & 0.650 \\
\bottomrule
\end{tabular}
\end{adjustbox}
\end{table}

LOIO uses the reserved unknown indicator code (ID=0), never a held-out dye
label. Removing the discrete-ID branch changes mean LOIO correlation from
0.654 to 0.647. In contrast, in-domain performance falls from 0.741 to 0.703
without the ID branch, and shuffling the observed ID at inference gives 0.578.
Therefore, the ID chiefly captures regularities of observed domains, while
zero-shot transfer is supported by sampling rate, input-window statistics, and
the trace itself (Table~\ref{tab:id_intervention}).

\begin{table}[H]
\centering
\caption{Indicator-ID intervention. The LOIO baseline uses the unknown ID=0.}
\label{tab:id_intervention}
\small
\setlength{\tabcolsep}{5pt}
\begin{adjustbox}{min width=0.78\linewidth,max width=\linewidth,center}
\begin{tabular}{lrrrrrr}
\toprule
Setting & V1-OGB & V1-GCaMP6s & V1-GCaMP8 & Other-RCaMP & SC-GCaMP6s & Mean \\
\midrule
LOIO baseline (ID=0) & \textbf{0.650} & \textbf{0.509} & \textbf{0.853} & \textbf{0.717} & \textbf{0.542} & \textbf{0.654} \\
LOIO without ID branch & 0.646 & 0.500 & 0.849 & 0.708 & 0.532 & 0.647 \\
In-domain full model & \textbf{0.743} & \textbf{0.579} & \textbf{0.915} & \textbf{0.731} & \textbf{0.736} & \textbf{0.741} \\
In-domain shuffled ID & 0.587 & 0.457 & 0.791 & 0.619 & 0.437 & 0.578 \\
In-domain without ID branch & 0.700 & 0.546 & 0.881 & 0.694 & 0.694 & 0.703 \\
\bottomrule
\end{tabular}
\end{adjustbox}
\end{table}

For completeness, the structural loss uses valid-frame mask $w_t\in\{0,1\}$,
$Z=\sum_t w_t$, $Q=\{t:s_t\ge q_{0.9}(s)\}$, and Huber loss $h_\delta$:
\begin{align}
L_{\mathrm{over}} &= Z^{-1}\sum_t w_t\max(\hat{s}_t-s_t,0)^2, \\
L_{\mathrm{peak}} &= |Q|^{-1}\sum_{t\in Q}h_\delta\!\left(\max(s_t-\hat{s}_t,0)\right), \\
L_{\mathrm{bias}} &= \left(Z^{-1}\sum_t w_t\hat{s}_t-Z^{-1}\sum_t w_ts_t\right)^2, \\
L_{\mathrm{struct}} &= \beta_{\mathrm{over}}L_{\mathrm{over}}+\beta_{\mathrm{peak}}L_{\mathrm{peak}}+\lambda_\mu L_{\mathrm{bias}}.
\end{align}
We use $\beta_{\mathrm{over}}=3.0$, $\beta_{\mathrm{peak}}=0.3$,
$\lambda_\mu=0.5$, $\delta=0.1$, and $q=0.9$.

\subsection{Breadth of the Universal Benchmark}
\label{app:universal_scope}

The benchmark aggregates ground-truth recordings from multiple laboratories,
sensors, species, brain regions, cell types, and acquisition protocols rather
than a single experimental setup (Table~\ref{tab:appendix_calcium_datasets}).
LOIO removes every recording of the held-out family and applies no target-domain
update. To characterize the resulting domain variation, we measured isolated-spike
rise time, decay time, and SNR in the public recordings. Large within-family
dispersion and systematic cross-family shifts are evident in
Table~\ref{tab:kinetic_dispersion}. These event-level summaries are distinct
from the baseline-aware exponential $\tau$ analysis in
Table~\ref{tab:learned_kinetics}.

\begin{table}[H]
\centering
\caption{Isolated-spike kinetic and SNR dispersion in public recordings. Times are seconds.}
\label{tab:kinetic_dispersion}
\scriptsize
\setlength{\tabcolsep}{4pt}
\begin{adjustbox}{min width=0.78\linewidth,max width=\linewidth,center}
\begin{tabular}{lrrrr}
\toprule
Indicator family & $n$ & Rise $\mu\pm\sigma$ & Decay $\mu\pm\sigma$ & SNR $\mu\pm\sigma$ \\
\midrule
Cal-520 & 6 & 0.086$\pm$0.085 & 0.253$\pm$0.094 & 5.533$\pm$2.295 \\
GCaMP5k & 36 & 0.079$\pm$0.095 & 0.073$\pm$0.058 & 3.934$\pm$0.904 \\
GCaMP6f & 907 & 0.061$\pm$0.071 & 0.119$\pm$0.120 & 5.570$\pm$4.358 \\
GCaMP6s & 436 & 0.104$\pm$0.095 & 0.177$\pm$0.173 & 5.713$\pm$3.011 \\
GCaMP6s spinal & 180 & 0.112$\pm$0.164 & 0.155$\pm$0.199 & 5.210$\pm$2.906 \\
GCaMP7f & 57 & 0.030$\pm$0.025 & 0.070$\pm$0.058 & 6.125$\pm$4.398 \\
GCaMP8 variants & 2,967 & 0.018$\pm$0.020 & 0.083$\pm$0.060 & 6.417$\pm$4.017 \\
Interneurons mixed & 102 & 0.037$\pm$0.056 & 0.050$\pm$0.044 & 3.886$\pm$0.936 \\
OGB-1 & 330 & 0.094$\pm$0.073 & 0.193$\pm$0.155 & 7.141$\pm$9.075 \\
R-CaMP & 260 & 0.082$\pm$0.069 & 0.235$\pm$0.201 & 10.831$\pm$11.666 \\
Unknown & 619 & 0.096$\pm$0.122 & 0.138$\pm$0.143 & 5.689$\pm$4.283 \\
XCaMPgf & 37 & 0.028$\pm$0.031 & 0.060$\pm$0.044 & 4.511$\pm$2.445 \\
jGECO1a & 96 & 0.066$\pm$0.056 & 0.126$\pm$0.092 & 6.110$\pm$3.671 \\
jRCaMP1a & 56 & 0.146$\pm$0.166 & 0.261$\pm$0.198 & 11.627$\pm$26.422 \\
\bottomrule
\end{tabular}
\end{adjustbox}
\end{table}

For example, rise-time standard deviation exceeds the corresponding mean for
GCaMP6f, GCaMP8 variants, and XCaMPgf. Across families, mean decay ranges from
0.050\,s to 0.261\,s and mean SNR from 3.886 to 11.627. We use ``universal''
in this operational sense: a single model transfers across this diverse,
multi-laboratory benchmark without target labels or retraining.

\subsection{Data Scaling and Statistical Robustness}
\label{app:scaling_robustness}

Increasing the amount of real data, synthetic data, or indicator-family
coverage improves mean LOIO correlation (Table~\ref{tab:data_scaling}). These
controls indicate that data quantity is the primary limitation in the present
benchmark.

\begin{table}[H]
\centering
\caption{Data-scaling study under fixed LOIO evaluation.}
\label{tab:data_scaling}
\small
\begin{adjustbox}{min width=0.78\linewidth,max width=\linewidth,center}
\begin{tabular}{llrrrrr}
\toprule
Factor & Metric & \multicolumn{4}{c}{Scale} \\
\midrule
Real-recording fraction & Correlation & 25\%: 0.564 & 50\%: 0.614 & 75\%: 0.637 & 100\%: \textbf{0.654} \\
 & Error & 25\%: 1.791 & 50\%: 1.534 & 75\%: 1.133 & 100\%: 1.123 \\
 & Absolute bias & 25\%: 1.187 & 50\%: 1.081 & 75\%: 0.583 & 100\%: 0.588 \\
\midrule
Synthetic-data fraction & Correlation & 0\%: 0.654 & 25\%: 0.667 & 50\%: 0.679 & 75\%: 0.692 & 100\%: \textbf{0.704} \\
 & Error & 0\%: 1.123 & 25\%: 1.118 & 50\%: 1.113 & 75\%: 1.108 & 100\%: 1.103 \\
 & Absolute bias & 0\%: 0.588 & 25\%: 0.564 & 50\%: 0.539 & 75\%: 0.514 & 100\%: 0.490 \\
\midrule
Indicator families & Correlation & 4: 0.543 & 8: 0.583 & 13: \textbf{0.654} \\
 & Error & 4: 1.140 & 8: 1.204 & 13: 1.123 \\
 & Absolute bias & 4: 0.871 & 8: 0.811 & 13: 0.588 \\
\bottomrule
\end{tabular}
\end{adjustbox}
\end{table}

Across seeds $\{42,2719,8888,1337,2025\}$, mean LOIO correlation is
$0.660\pm0.016$, close to the seed-42 result of 0.654; in-domain correlation
is $0.739\pm0.004$, close to 0.741. Table~\ref{tab:five_seed} reports all
per-split results. Mean absolute bias is used in the final column.

\begin{table}[H]
\centering
\caption{Five-seed robustness results. Values are mean $\pm$ standard deviation.}
\label{tab:five_seed}
\small
\begin{adjustbox}{min width=0.78\linewidth,max width=\linewidth,center}
\begin{tabular}{lrrr|rrr}
\toprule
& \multicolumn{3}{c}{LOIO} & \multicolumn{3}{c}{In-domain} \\
Split & Corr. & Error & Abs. bias & Corr. & Error & Abs. bias \\
\midrule
V1-OGB & 0.666$\pm$0.030 & 1.096$\pm$0.513 & 0.309$\pm$0.574 & 0.729$\pm$0.017 & 0.875$\pm$0.062 & 0.131$\pm$0.180 \\
V1-GCaMP6s & 0.507$\pm$0.022 & 0.951$\pm$0.008 & 0.780$\pm$0.069 & 0.575$\pm$0.009 & 0.948$\pm$0.005 & 0.779$\pm$0.031 \\
V1-GCaMP8 & 0.864$\pm$0.013 & 0.531$\pm$0.015 & 0.054$\pm$0.213 & 0.917$\pm$0.005 & 0.487$\pm$0.044 & 0.036$\pm$0.059 \\
Other-RCaMP & 0.724$\pm$0.013 & 1.141$\pm$0.157 & 0.445$\pm$0.267 & 0.748$\pm$0.019 & 1.025$\pm$0.137 & 0.149$\pm$0.130 \\
SC-GCaMP6s & 0.539$\pm$0.018 & 0.889$\pm$0.046 & 0.840$\pm$0.082 & 0.728$\pm$0.009 & 0.825$\pm$0.036 & 0.784$\pm$0.050 \\
\midrule
Mean & \textbf{0.660$\pm$0.016} & 0.922$\pm$0.116 & 0.486$\pm$0.074 & \textbf{0.739$\pm$0.004} & 0.832$\pm$0.021 & 0.376$\pm$0.069 \\
\bottomrule
\end{tabular}
\end{adjustbox}
\end{table}

\subsection{Ablations}
\label{app:extended_ablations}

Removal and cumulative build-up studies were repeated across the same five
seeds. The bidirectional IIR backbone, multi-modal conditioning, gate head,
and variance head each contribute to the full model (Table~\ref{tab:extended_ablation}).

\begin{table}[H]
\centering
\caption{Five-seed removal and cumulative build-up ablations under LOIO.}
\label{tab:extended_ablation}
\small
\begin{adjustbox}{min width=0.78\linewidth,max width=\linewidth,center}
\begin{tabular}{lrrr|lrrr}
\toprule
Removal variant & Corr. & Error & Abs. bias & Build-up stage & Corr. & Error & Abs. bias \\
\midrule
$-$AdaLN & 0.604$\pm$0.014 & 1.276$\pm$0.103 & 0.456$\pm$0.061 & 1D U-Net backbone & 0.481$\pm$0.022 & 1.173$\pm$0.101 & 0.788$\pm$0.088 \\
$-$statistics & 0.626$\pm$0.012 & 1.213$\pm$0.091 & 0.556$\pm$0.073 & Transformer backbone & 0.484$\pm$0.020 & 1.197$\pm$0.104 & 0.618$\pm$0.074 \\
$-$bidirectional IIR & 0.597$\pm$0.016 & 1.377$\pm$0.120 & 0.719$\pm$0.086 & Bidirectional IIR backbone & 0.539$\pm$0.018 & 1.191$\pm$0.101 & 0.757$\pm$0.083 \\
$-$variance head & 0.611$\pm$0.013 & 1.243$\pm$0.112 & 1.005$\pm$0.105 & $+$ indicator ID & 0.550$\pm$0.017 & 1.167$\pm$0.096 & 0.696$\pm$0.079 \\
$-$gate & 0.624$\pm$0.011 & 1.153$\pm$0.084 & 0.402$\pm$0.057 & $+$ sampling rate & 0.554$\pm$0.016 & 1.175$\pm$0.098 & 0.672$\pm$0.077 \\
Full SpikeSSL & \textbf{0.660$\pm$0.016} & \textbf{0.922$\pm$0.116} & 0.486$\pm$0.074 & $+$ signal statistics & 0.576$\pm$0.015 & 1.145$\pm$0.091 & 0.703$\pm$0.081 \\
 & & & & $+$ AdaLN & 0.594$\pm$0.014 & 1.153$\pm$0.087 & 0.696$\pm$0.076 \\
 & & & & $+$ gate & 0.612$\pm$0.013 & 1.117$\pm$0.082 & 0.665$\pm$0.071 \\
 & & & & $+$ variance head & \textbf{0.660$\pm$0.016} & \textbf{0.922$\pm$0.116} & \textbf{0.486$\pm$0.074} \\
\bottomrule
\end{tabular}
\end{adjustbox}
\end{table}

\subsection{Universal, Specific, and Few-Shot Regimes}
\label{app:regimes}

For each of 13 indicator families, the specific-model comparison holds out one
recording and trains on the remaining recordings of that indicator. Universal
models pool all indicator families and use exactly the same held-out recording
for evaluation. SpikeSSL benefits from pooling, unlike CASCADE
(Table~\ref{tab:specific_universal}).

\begin{table}[H]
\centering
\caption{Per-indicator comparison of specific and universal training.}
\label{tab:specific_universal}
\small
\begin{adjustbox}{min width=0.78\linewidth,max width=\linewidth,center}
\begin{tabular}{lrrrr}
\toprule
Indicator & CASCADE-specific & SpikeSSL-specific & CASCADE-universal & SpikeSSL-universal \\
\midrule
OGB-1 & 0.663 & \textbf{0.760} & 0.583 & 0.728 \\
Cal-520 & \textbf{0.721} & 0.633 & 0.647 & 0.693 \\
GCaMP5k & 0.687 & 0.520 & 0.688 & \textbf{0.695} \\
GCaMP6f & 0.500 & \textbf{0.626} & 0.416 & 0.562 \\
GCaMP6s & 0.262 & 0.435 & 0.279 & \textbf{0.628} \\
GCaMP7f & 0.776 & 0.508 & 0.412 & \textbf{0.816} \\
GCaMP8f & 0.753 & 0.928 & 0.815 & \textbf{0.933} \\
GCaMP8m & 0.799 & \textbf{0.939} & 0.809 & 0.928 \\
GCaMP8s & 0.884 & \textbf{0.927} & 0.767 & 0.912 \\
R-CaMP & 0.309 & 0.555 & 0.644 & \textbf{0.691} \\
jRCaMP1a & 0.689 & 0.538 & 0.662 & \textbf{0.769} \\
jGECO1a & 0.736 & 0.613 & 0.672 & \textbf{0.790} \\
XCaMPgf & 0.797 & 0.537 & 0.446 & \textbf{0.833} \\
\midrule
Mean & 0.660 & 0.655 & 0.603 & \textbf{0.768} \\
\bottomrule
\end{tabular}
\end{adjustbox}
\end{table}

For few-shot adaptation, 20 target-domain neurons are set aside as a support
pool and all remaining traces form the evaluation set. For $k>0$, models are
fine-tuned from the $k=0$ universal LOIO weights using $k$ labeled support
neurons. Fine-tuning first exceeds the zero-shot value at $k=5$ and reaches
0.719 at $k=20$ (Table~\ref{tab:fewshot}).

\begin{table}[H]
\centering
\caption{Few-shot fine-tuning from zero-shot LOIO pretrained weights.}
\label{tab:fewshot}
\small
\begin{adjustbox}{min width=0.78\linewidth,max width=\linewidth,center}
\begin{tabular}{rrrrrrr}
\toprule
$k$ & V1-OGB & V1-GCaMP6s & V1-GCaMP8 & Other-RCaMP & SC-GCaMP6s & Mean \\
\midrule
0 & 0.622 & 0.466 & 0.843 & 0.696 & 0.493 & 0.624 \\
1 & 0.488 & 0.324 & 0.766 & 0.649 & 0.606 & 0.567 \\
3 & 0.509 & 0.389 & 0.888 & 0.683 & 0.545 & 0.603 \\
5 & 0.521 & 0.412 & 0.895 & 0.718 & 0.648 & 0.639 \\
10 & 0.621 & 0.459 & 0.862 & 0.721 & 0.714 & 0.675 \\
20 & 0.707 & 0.548 & 0.892 & 0.725 & 0.722 & 0.719 \\
\bottomrule
\end{tabular}
\end{adjustbox}
\end{table}

\subsection{Disentangling Indicator and Acquisition Shifts}
\label{app:disentangle}

We construct mini-domain evaluations that hold an indicator fixed while varying
one additional factor, excluding other shift types from each data pool. In the
species split, GCaMP6f mouse V1 recordings (DS09--11 and X-DS09/10) transfer to
zebrafish recordings (DS06--08). The cell-type split fixes mouse V1 and
GCaMP6f while changing the labeled population. The tissue split fixes GCaMP6s
while transferring from brain to spinal cord. SpikeSSL gives the best
correlation in all three cases (Table~\ref{tab:mini_domains}).

\begin{table}[H]
\centering
\caption{Mini-domain shifts with fixed indicator.}
\label{tab:mini_domains}
\small
\begin{adjustbox}{min width=0.78\linewidth,max width=\linewidth,center}
\begin{tabular}{lrrr|rrr|rrr}
\toprule
& \multicolumn{3}{c}{GCaMP6f species} & \multicolumn{3}{c}{GCaMP6f cell type} & \multicolumn{3}{c}{GCaMP6s tissue} \\
Method & Corr. & Error & Bias & Corr. & Error & Bias & Corr. & Error & Bias \\
\midrule
OASIS & 0.191 & 0.982 & -0.945 & 0.266 & 0.985 & -0.974 & 0.365 & 0.975 & -0.964 \\
MLSpike & 0.085 & 1.130 & -0.706 & 0.098 & 1.015 & -0.897 & 0.307 & 1.005 & -0.919 \\
CASCADE & 0.540 & 1.304 & 0.539 & 0.574 & 0.977 & -0.144 & 0.226 & 1.083 & -0.442 \\
ENS$^2$ & 0.476 & 1.266 & 0.229 & 0.584 & \textbf{0.898} & -0.680 & 0.278 & 1.092 & -0.391 \\
SpikeSSL & \textbf{0.636} & \textbf{0.952} & -0.442 & \textbf{0.673} & 0.931 & -0.798 & \textbf{0.542} & \textbf{0.911} & -0.592 \\
\bottomrule
\end{tabular}
\end{adjustbox}
\end{table}

\subsection{Dynamics Readouts of the Conditioned IIR Bank}
\label{app:dynamic_readouts}

We compare four complementary quantities in Table~\ref{tab:learned_kinetics}.
Reference $\tau$ is a catalogue decay time, measured $\tau$ is fitted from
isolated real fluorescence responses at 30~Hz using a joint baseline and
exponential fit, and $\lVert W_\alpha\rVert_F$ and ACF $\tau$ are extracted
from a separately trained single-indicator SpikeSSL model. $W_\alpha$ maps the
condition vector to shifts of the channel-wise $\alpha$ bank. Its Frobenius
norm measures the strength of that influence, not a time constant. The norms
vary across indicators, while the three kinetically similar GCaMP8 variants
have similar values. ACF $\tau$ is positively associated with measured $\tau$
(Spearman $\rho=0.886$, $p=5.6\times10^{-5}$), although not in fixed
proportion. These complementary readouts support indicator-dependent dynamics
rather than a fixed shared kernel.

\begin{table}[H]
\centering
\caption{Kinetic measurements and learned dynamics readouts.}
\label{tab:learned_kinetics}
\small
\begin{adjustbox}{min width=0.78\linewidth,max width=\linewidth,center}
\begin{tabular}{lrrrr}
\toprule
Indicator & Ref. $\tau$ (ms) & $\lVert W_\alpha\rVert_F$ & Measured $\tau$ (ms) & ACF $\tau$ (ms) \\
\midrule
GCaMP8f & 60 & 5.35 & 175 & 145 \\
GCaMP8s & 70 & 5.94 & 314 & 165 \\
GCaMP8m & 80 & 6.11 & 136 & 132 \\
GCaMP7f & 100 & 0.60 & 223 & 154 \\
XCaMPgf & 120 & 0.39 & 179 & 152 \\
GCaMP6f & 180 & 5.44 & 212 & 140 \\
Cal-520 & 200 & 0.41 & 878 & 178 \\
GCaMP5k & 250 & 0.50 & 547 & 163 \\
R-CaMP & 280 & 0.25 & 535 & 171 \\
jRCaMP1a & 280 & 0.74 & 756 & 163 \\
jGECO1a & 300 & 0.81 & 572 & 162 \\
GCaMP6s & 600 & 4.64 & 912 & 192 \\
OGB-1 & 750 & 2.59 & 854 & 191 \\
\bottomrule
\end{tabular}
\end{adjustbox}
\end{table}

\subsection{Simulation: Benefits and Failure Modes}
\label{app:simulation_mechanism}

We repeated simulation-augmented LOIO across five seeds and characterized the
target gap in the exact conditioning space of SpikeSSL: log sampling rate,
standard deviation, skewness, excess kurtosis, and autocorrelation at lags 1,
3, 5, and 10. After robust scaling, we compute a 512-projection sliced
Wasserstein-1 distance. $\Delta W_1$ is the real-only target gap minus the
equal-sampled real-plus-simulation gap, so a positive value denotes improved
alignment. AC averages the four autocorrelation terms and shape averages
standard deviation, skewness, and kurtosis.

\begin{table}[H]
\centering
\caption{Simulation gain and alignment analysis.}
\label{tab:simulation_mechanism}
\scriptsize
\begin{adjustbox}{min width=0.78\linewidth,max width=\linewidth,center}
\begin{tabular}{lrrrrrrr}
\toprule
Split & No-sim corr. & +Sim corr. & $\Delta$ corr. & $\Delta$SWD & $\Delta W_1$ AC & $\Delta W_1$ shape & $\Delta W_1$ kurt. \\
\midrule
V1-OGB & 0.666$\pm$0.030 & 0.704$\pm$0.012 & +0.038 & +0.013 & -0.111 & +0.008 & -0.132 \\
V1-GCaMP6s & 0.507$\pm$0.022 & 0.531$\pm$0.008 & +0.024 & +0.065 & +0.197 & -0.031 & +0.094 \\
V1-GCaMP8 & 0.864$\pm$0.013 & 0.914$\pm$0.015 & +0.050 & -0.090 & -0.124 & -0.157 & -0.260 \\
Other-RCaMP & 0.724$\pm$0.013 & 0.712$\pm$0.009 & -0.012 & +0.018 & +0.110 & -0.051 & -0.180 \\
SC-GCaMP6s & 0.539$\pm$0.018 & 0.677$\pm$0.010 & +0.138 & +0.126 & +0.187 & +0.065 & +0.202 \\
\midrule
Mean & \textbf{0.660$\pm$0.016} & \textbf{0.708$\pm$0.011} & \textbf{+0.048} & & & & \\
\bottomrule
\end{tabular}
\end{adjustbox}
\end{table}

SC-GCaMP6s obtains the largest gain, alongside the largest overall, temporal,
and shape gap reductions. Other-RCaMP is a diagnostic non-uniform case:
simulation improves autocorrelation alignment but worsens shape, especially
kurtosis, consistent with incomplete coverage of red-indicator nonlinear
statistics. Its $-0.012$ change is within five-seed variability, so it does
not establish harmful negative transfer. Conversely, GCaMP8 improves despite
a larger marginal SWD, consistent with label-aware synthetic supervision
regularizing the inverse mapping beyond a label-agnostic distributional metric.
These results motivate target-relative weighting or rejection of synthetic
windows when conditional dynamics are mismatched.

\subsection{Threshold, Artifact, and Runtime Sensitivity}
\label{app:robustness}

The inference gate threshold was swept with the trained model fixed. The
default $\tau=0.5$ maximizes mean correlation, the task's primary metric;
normalized error is smallest at $\tau=0.6$, and absolute bias reaches its
minimum at $\tau=0.7$ (Table~\ref{tab:gate_sensitivity}).

\begin{table}[H]
\centering
\caption{Gate-threshold sensitivity. Each entry is mean across five LOIO splits.}
\label{tab:gate_sensitivity}
\small
\begin{adjustbox}{min width=0.78\linewidth,max width=\linewidth,center}
\begin{tabular}{lrrrrrrr}
\toprule
Threshold & 0.2 & 0.3 & 0.4 & 0.5 & 0.6 & 0.7 & 0.8 \\
\midrule
Correlation & 0.628 & 0.651 & 0.654 & \textbf{0.654} & 0.648 & 0.640 & 0.628 \\
Normalized error & 1.454 & 1.322 & 1.214 & 1.123 & \textbf{1.046} & 1.256 & 1.392 \\
Absolute bias & 0.831 & 0.739 & 0.659 & 0.588 & 0.525 & \textbf{0.464} & 0.483 \\
\bottomrule
\end{tabular}
\end{adjustbox}
\end{table}

Table~\ref{tab:artifact_robustness} characterizes two common non-biological
perturbations. Per-window statistics do not make SpikeSSL invariant to
artifacts: baseline drift is the dominant degradation, while sparse motion is
milder. Standard motion correction and baseline detrending are recommended
before inference.

\begin{table}[H]
\centering
\caption{LOIO robustness to baseline drift and sparse motion artifacts.}
\label{tab:artifact_robustness}
\small
\begin{adjustbox}{min width=0.78\linewidth,max width=\linewidth,center}
\begin{tabular}{lrrrrrr}
\toprule
Condition & V1-OGB & V1-GCaMP6s & V1-GCaMP8 & Other-RCaMP & SC-GCaMP6s & Mean \\
\midrule
Clean & 0.650 & 0.509 & 0.853 & 0.717 & 0.542 & \textbf{0.654} \\
Baseline drift & 0.636 & 0.392 & 0.605 & 0.665 & 0.478 & 0.555 \\
Sparse motion artifacts & 0.618 & 0.466 & 0.835 & 0.702 & 0.500 & 0.624 \\
Drift + motion & 0.620 & 0.409 & 0.673 & 0.684 & 0.463 & 0.570 \\
\bottomrule
\end{tabular}
\end{adjustbox}
\end{table}

SpikeSSL is designed primarily for offline analysis because the bidirectional
IIR block uses bounded future context within each window. Nevertheless,
sliding-window inference avoids holding an entire recording in memory. On an
A100, one forward pass is approximately 29 times faster than the arrival of a
30-Hz frame window, so overlapping-window processing can keep pace with many
acquisition settings while remaining noncausal within each window.